\pdfoutput=1

\documentclass[11pt]{article}

\usepackage[]{acl}

\usepackage{times}
\usepackage{latexsym}
\usepackage{hyperref}
\usepackage{url}
\usepackage{authblk}
\usepackage{graphicx}
\usepackage{algorithm}
\usepackage{algpseudocode}
\usepackage{multirow}
\usepackage{array}
\usepackage{booktabs}
\usepackage{adjustbox}
\usepackage{wrapfig}
\usepackage{mathtools}

\definecolor{alizarin}{rgb}{0.82, 0.1, 0.26}
\definecolor{amaranth}{rgb}{0.9, 0.17, 0.31}
\definecolor{americanrose}{rgb}{1.0, 0.01, 0.24}
\definecolor{awesome}{rgb}{1.0, 0.13, 0.32}
\definecolor{darkterracotta}{rgb}{0.8, 0.31, 0.36}

\usepackage[T1]{fontenc}

\usepackage[utf8]{inputenc}

\usepackage{microtype}

\usepackage{inconsolata}

\title{Self-Specialization:\\ Uncovering Latent Expertise within Large Language Models}

\author[1,3]{\textbf{Junmo Kang}\thanks{Work done during internship at MIT-IBM Watson AI Lab.}}
\author[2]{\textbf{Hongyin Luo}}
\author[3]{\textbf{Yada Zhu}}
\author[2]{\textbf{Jacob Hansen}}
\author[2]{\\\textbf{James Glass}}
\author[3]{\textbf{David Cox}}
\author[1]{\\\textbf{Alan Ritter}}
\author[3]{\textbf{Rogerio Feris}}
\author[3]{\textbf{Leonid Karlinsky}}
\affil[1]{Georgia Institute of Technology}
\affil[2]{Massachusetts Institute of Technology}
\affil[3]{MIT-IBM Watson AI Lab}
\affil[ ]{\texttt{\textmd{\small{junmo.kang@gatech.edu}}}}

\begin{document}
\maketitle
\begin{abstract}
Recent works have demonstrated the effectiveness of self-alignment in which a large language model is aligned to follow general instructions using instructional data generated from the model itself starting from a handful of human-written seeds. 
Instead of general alignment, in this work, we focus on self-alignment for expert domain specialization (e.g., biomedicine, finance).
As a preliminary, we quantitively show the marginal effect that generic instruction-following training has on downstream expert domains' performance. To remedy this, we propose \textbf{self-specialization} - allowing for effective model specialization while achieving cross-task generalization by leveraging only a few labeled seeds.
Self-specialization offers a data- and parameter-efficient
way of ``carving out'' an expert model out of a generalist pre-trained LLM. 
Exploring a variety of popular open large models as a base for specialization, our experimental results in both biomedical and financial domains show that our self-specialized models outperform their base models by a large margin, and even larger models that are generally instruction-tuned or that have been adapted to the target domain by other means.
\end{abstract}

\section{Introduction}

Instruction-tuning \citep{ouyang2022training, wei2022finetuned, naturalinstructions, su2022one} of large language models (LLMs) offers a mechanism to adeptly guide models using specific directives, thereby enhancing their versatility across diverse tasks. However, as promising as this concept might seem, it poses an inherent challenge: the substantial need for quality data \citep{chung2022scaling, Wan2023Poisoning, kpf2023openassistant}. The very premise of instruction-tuning hinges on the availability of well-crafted, human-annotated data, a resource that is both time-consuming and challenging to scale efficiently \citep{honovich2022unnatural, kang-etal-2023-distill}. 

When it comes to specialized domains, such as biomedicine, it is more challenging to acquire human labels, due to the need for expert annotators
\citep{wang2023empower}.
While adaptation through in-domain pre-training \citep{gururangan-etal-2020-dont, wu2023pmcllama} has been shown to be 
effective, this approach requires extensive (unlabeled) target-domain data, in addition to significant computational resources.  Moreover, prior work has shown the benefits of adaptive pre-training can be less than those achieved by moderate amounts of fine-tuning data from the target domain \citep{bai-etal-2021-pre}.

\begin{figure}[!t]
    \centering
    \includegraphics[width=0.9\linewidth]{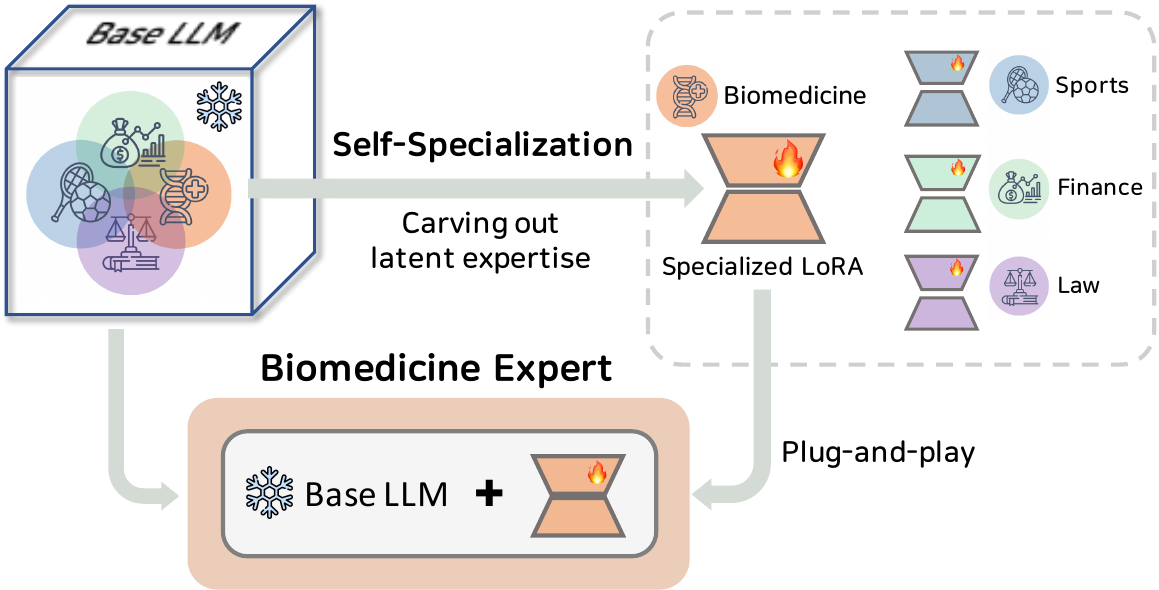}
    \caption{Self-specialization concept. Expertise in various domains is mixed and latent within base LLMs, and can be carved out through self-specialization.}
    \label{fig:concept}
\end{figure}

Emerging as a promising solution to this data-intensive challenge in the context of instruction-tuning is the approach of self-alignment \citep{wang2022self, sun2023principledriven}. By allowing LLMs to automatically generate instructional data from minimal human-authored seeds, self-alignment presents a means to harness the internal general knowledge of models, which results from extensive pre-training on internet corpora \citep{devlin-etal-2019-bert, 2020t5, NEURIPS2020_1457c0d6}, without extensive human annotations.

However, a pertinent question remains: How effective are the self-aligned models when applied to more niche domains, such as biomedicine?
Given that both the initial pre-training and subsequent self-alignment are general, the knowledge embedded in LLM parameters may be a mixture of semantics and various domains. This raises questions about their effectiveness in specialized domains, despite the aims of instruction-tuning and self-alignment for cross-task generalization.
In our preliminary study, however, we find that existing models such as Alpaca \citep{alpaca} and Dromedary \citep{sun2023principledriven}, although aligned, exhibit only a modest degree of improvement 
within the specialized domains. These observations underline the need for focused approaches that can leverage the domain expertise existing in the base models, to ensure the self-generated instruction-tuning data remains both contextually appropriate and accurate.

In this work, we explore the possibility of \textbf{self-specialization} (Fig. \ref{fig:concept}). Drawing inspiration from the foundational principles of self-alignment, self-specialization goes a step further by incorporating domain-specific seed instructions and is further bolstered by parameter-efficient fine-tuning, as well as optional iterative refinement and retrieval components. Our goal is to guide models beyond generic alignment, directing them to generate data that are not just contextually fitting for a specialized domain but also maintain high accuracy.

We evaluate our self-specialized models within the biomedical and finance domains (20 datasets in total), and across a variety of base models that we specialize. Surprisingly, despite the simplicity of our approach, our results present a compelling case for self-specialization significantly outperforming the base models, and even larger models that are generally instruction-tuned or specifically pre-trained on the target domain. Notably, our self-specialized one based on MPT-30B \citep{MosaicML2023Introducing} for biomedicine even surpasses larger models (based on LLaMA-65B \citep{touvron2023llama}), including the ones improved through self-alignment by leading methods \citep{wang2022self, sun2023principledriven}. 
%

\section{Preliminaries: Benchmarking Existing Aligned Models}
\label{sec:prelim}

To motivate our exploration of self-specialization, we first begin by addressing a fundamental question: How well do generally aligned models perform on specialized domains? While 
popular models, such as Alpaca \citep{alpaca} and Dromedary \citep{sun2023principledriven}, have demonstrated effectiveness in following general instructions, it remains unclear whether general alignment can also elicit expertise for a certain domain. 
\begin{table}[t!]
\begin{center}
\begin{adjustbox}{width=0.45\textwidth}
\begin{tabular}{cccc}
\toprule
& \textsc{Base} & \multicolumn{2}{c}{\textsc{Aligned}} \\
\cmidrule(lr){2-2}\cmidrule(lr){3-4}
  Model & \textbf{LLaMA-65B} & \textbf{Alpaca-65B} & \textbf{Dromedary-65B} \\
  \midrule\midrule
Averaged & \multirow{2}{*}{43.87} & 46.39 & 45.10 \\
\textsc{\small{$F_1$-Score}} & & (\texttt{+2.52}) & (\texttt{+1.23}) \\
\bottomrule
\end{tabular}
\end{adjustbox}
\end{center}
\caption{Benchmarking results of a base LLaMA-65B and its aligned variants in a biomedical domain. The evaluation covers various NLP tasks such as question answering, information extraction, and classification. 5-shot results averaged across 10 datasets are presented.
}
\label{table:65b-benchmark}
\end{table}

Investigating this, we assess the capabilities of Alpaca and Dromedary against their base model, LLaMA-65B \citep{touvron2023llama}, on a collection of benchmarks within the biomedical domain. 
We evaluate Alpaca as an upper bound, due to its reliance on GPT-3.5-generated datasets \citep{ouyang2022training} via the self-instruct process \citep{wang2022self}, unlike Dromedary, which generates instructional data from its base model.
We use 10 biomedical NLP datasets (see Section \ref{subsec:setups} for details), covering a diverse set of tasks to ensure a comprehensive mix of content and also to look at the cross-task generalization, the core of instruction-tuning. 
Table \ref{table:65b-benchmark} summarizes the result.

We find that both Alpaca and Dromedary have only a slight (1.2 - 2.5) advantage over LLaMA in biomedicine. While they are aligned to handle a broad set of instructions, they do not seem to effectively improve their specialized domain expertise; intuitively trading their expertise for generality given finite parameters. In light of these findings, it becomes evident that for cases where we are only interested in expert domains for all our downstream tasks, 
there remains a large potential for improvement beyond the generic alignment.
This underscores the need for a model or approach, like self-specialization, that could potentially uncover specialization while maintaining cross-task generalizability with minimal supervision. 

\begin{figure*}[t!]
    \centering
    \includegraphics[width=0.95\textwidth]
    {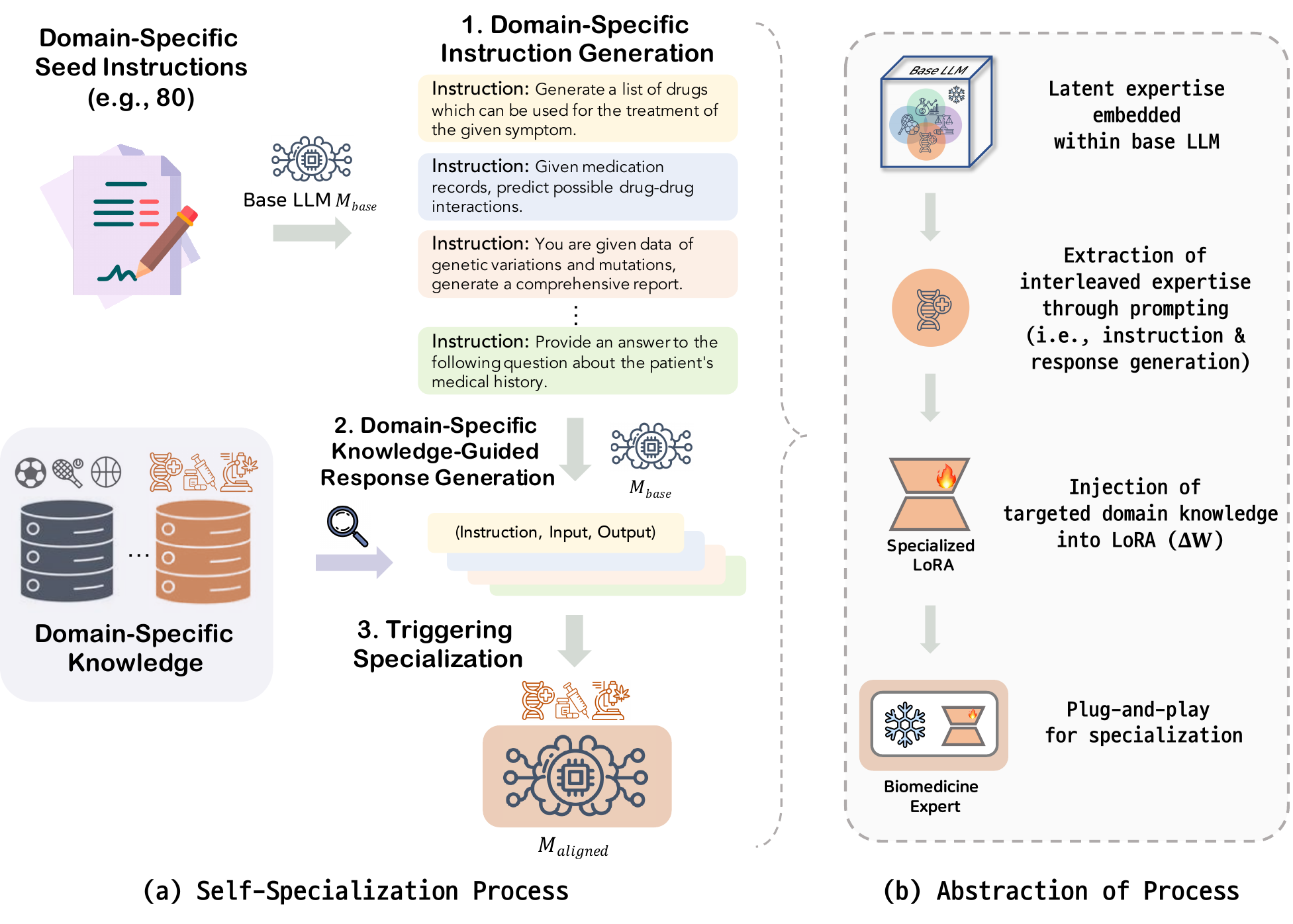}
    \caption{\textbf{Self-Specialization} overview. (a) We start with a small set of human-authored domain-specific seed instructions. The base model crafts synthetic instructions and corresponding input contexts tailored to that particular domain. Subsequently, during the response generation phase, responses are curated given the generated instruction and input pairs, optionally enhanced by infusing domain-relevant knowledge obtained via a retrieval component or iterative re-generation via our previous self-specialized model. Finally, in the specialization phase, the base model is tuned for specialization (w/ QLoRA) to uncover its target domain expertise. (b) Conceptually speaking, this process can be described as uncovering latent expertise within LLMs.}
    \label{fig:overview}
\end{figure*}

\section{Self-Specialization}
In this section, we describe our method called self-specialization illustrated in Figure \ref{fig:overview}. 

\subsection{Seed Demonstrations}
\label{subsec:seed}
Initially, we utilize a curated set of seed demonstrations $S$, consisting of a triplet $(i, c, y)$, comprised of instruction $i$, a context $c$ (e.g., passage), and a response $y$, respectively. Recognizing the difficulty of acquiring domain-specific data in real-world scenarios \citep{bai-etal-2021-pre}, we aim for a very minimal number of seeds: only 80 for the biomedical domain and 90 for the financial domain\footnote{While manual annotation of seed data is an assumed prerequisite for this initial step in self-alignment, we consider those numbers to be reasonable to annotate.}. We leverage established datasets such as Box \citep{parmar2022boxbart} for seed construction to fairly ensure quality (detailed in Section \ref{subsec:setups}). These seeds capture essential domain concepts but are insufficient to cover the entirety of domain knowledge. 
We posit that domain-relevant information, intermingled with the vast array of other information acquired during pre-training, can be effectively accessed and better utilized through our self-specialization approach, enabling these models to enhance their performance in specialized domains.
Seeds provide the primary scaffold upon which subsequent domain-specific instructions are built.

\subsection{Domain-Specific Instruction Generation}
\label{subsec:inst}
With the seed instructions in place, we move to generating domain-specific instructions. While these new instructions are grounded in the initial seeds, they grow to cover a comprehensive scope of the domain. Specifically, a base model $M_{base}$, such as MPT-30B \citep{MosaicML2023Introducing} which is large enough, is prompted to produce new combinations of $(i, c)$ given a handful of seed demonstrations which are randomly sampled from the initial seeds pool. The newly formed instructions $i$, coupled with their corresponding input contexts $c$, shape a blueprint that the model utilizes in the following stages.

\subsection{Domain-Specific Response Generation}
\label{subsec:resp}
In this phase, it is crucial for the responses not only to be correct but also to be well-aligned with the target domain. Intuitively, as this phase is conditioned on domain-specific instructions $\{i\}$ and corresponding contexts $\{c\}$, derived from domain-specific seeds, it may be sufficient to rely on the base model itself to generate domain-specific responses.
As an additional effort, we explore whether leveraging external domain-relevant knowledge would be beneficial for this case, inspired by \citet{frisoni-etal-2022-bioreader}. 
Therefore, we optionally allow $M_{base}$ to incorporate external knowledge via a retrieval component $M_{ret}$.
Specifically, forming the query $x$ as a concatenation of $i$ and $c$, ${M_{ret}}$ fetches top-$k$ relevant documents $d_{1:k}$.
\begin{equation*}
\small{
    d_{1:k} = M_{ret}(x = i \oplus c)
}
\end{equation*}

Then, each document $d_j$ is independently paired with the query $x$ to form a prompt to $M_{base}$, and the final domain-specific responses $y$ are produced from the final distribution computed by marginalizing over the probabilities of each of these $k$-combinations at each generation step. 
\begin{equation*}
\small{
\begin{multlined}
    p(y|x) = \\
    \prod_i^t \sum_j^k p_{ret}(d_j|x; M_{ret})\ p_{lm}(y_i|x, d_j, y_{1:i-1}; M_{base})
\label{eq:margin}
\end{multlined}
}
\end{equation*}
where $p_{ret}$ is a relevance score (similarity) from a retriever module and $p_{lm}$ represents the language model distribution.
By integrating such external information, while domain-relevant knowledge is deemed latent within LLMs, this step further encourages the generated target responses to be more nuanced and domain-specific, leading to additional improvements (Section \ref{subsec:analyses}).

\subsection{Triggering Specialization}
\label{subsec:trig}
Upon establishing a set of domain-specific instructions/responses, $M_{base}$ undergoes tuning using the self-generated data, adjusting its internal parameters
to cater specifically to the domain's nuances. This step is crucial, marking the model's transformation from being generally competent to being domain-specialized while preserving cross-task generalizability, thus resulting in the final self-aligned domain-specialized model: $M_{aligned}$.

\subsection{Iterative Self-Specialization}
\label{subsec:iter}
In the spirit of continuous improvement, our approach optionally supports iterative self-specialization
via re-generating
instructions and responses with the better-aligned model $M_{aligned}$. 
This process has the potential of refining the model's domain expertise with each iteration (of considering the previous iteration $M_{aligned}$ as base each time), iteratively improving its responses. 

\begin{table*}[t!]
\centering
\begin{adjustbox}{width=0.9\textwidth}
\begin{tabular}{llcccccc}
\toprule
\multicolumn{2}{c}{\textsc{Biomedicine}} &  
\multicolumn{2}{c}{\textit{k=0}} & \multicolumn{2}{c}{\textit{k=1}} & \multicolumn{2}{c}{\textit{k=5}} \\ 
\cmidrule(lr){3-4}\cmidrule(lr){5-6}\cmidrule(lr){7-8}
 \textbf{Task} & \textbf{Dataset} & \textbf{Base} & \textbf{Self-Specialized} & \textbf{Base} & \textbf{Self-Specialized} & \textbf{Base} & \textbf{Self-Specialized} \\ 
\midrule\midrule
\multirow{4}{*}{\shortstack{QA}} 
 & BioASQ-Factoid & 30.90 & \textbf{37.35} & 47.56 & \textbf{55.04} & 51.96 & \textbf{57.61} \\
 & BioASQ-List & 46.06 & \textbf{46.99} & \textbf{47.57} & 44.55 & 35.09 & \textbf{42.17} \\
 & BioASQ-Yesno\hyperlink{bioasq-footnote}{$^{3}$} & 21.20 & \textbf{85.27} & 10.80 & \textbf{94.00} & \ \ 8.80 & \textbf{95.20} \\
 & PubMedQA & 11.98 & \textbf{24.16} & \textbf{28.89} & 24.87 & \textbf{31.69} & 31.31 \\
\cmidrule(lr){1-8}
\multirow{3}{*}{\shortstack{NER}} 
 & AnatEM & \ \ 9.63 & \textbf{11.99} & \ \ 7.57 & \textbf{15.76} & \ \ 6.59 & \textbf{21.25} \\
 & BioNLP13CG & 24.79 & \textbf{24.93} & 21.76 & \textbf{31.80} & 26.03 & \textbf{41.16} \\
 & NCBI & \textbf{18.46} & 14.35 & 27.88 & \textbf{43.11} & 17.99 & \textbf{46.54} \\
\cmidrule(lr){1-8}
\multirow{1}{*}{\shortstack{RE}} 
 & DDI & \textbf{51.00} & 49.40 & 49.20 & \textbf{51.60} & 49.38 & \textbf{53.40} \\
\cmidrule(lr){1-8}
\multirow{1}{*}{\shortstack{SA}} 
 & Medical Drugs & 35.00 & \textbf{65.80} & 11.40 & \textbf{54.60} & 11.40 & \textbf{32.80} \\
\cmidrule(lr){1-8}
\multirow{1}{*}{\shortstack{DC}} 
 & HoC & \ \ 2.44 & \textbf{\ \ 6.01} & \textbf{13.91} & \ \ 7.61 & \textbf{62.84} & 62.65 \\
\midrule
\multicolumn{2}{c}{Average\qquad\qquad\qquad} & 25.15 & \textbf{36.63} & 26.65 & \textbf{42.29} & 30.18 & \textbf{48.41} \\
\bottomrule
\end{tabular}
\end{adjustbox}
\smallskip

\begin{adjustbox}{width=0.9\textwidth}
\begin{tabular}{llcccccc}
\toprule
\multicolumn{2}{c}{\textsc{Finance}\qquad\qquad} &  
\multicolumn{2}{c}{\textit{k=0}} & \multicolumn{2}{c}{\textit{k=1}} & \multicolumn{2}{c}{\textit{k=5}} \\  
\cmidrule(lr){3-4}\cmidrule(lr){5-6}\cmidrule(lr){7-8}
 \textbf{Task} & \textbf{Dataset} & \textbf{Base} & \textbf{Self-Specialized} & \textbf{Base} & \textbf{Self-Specialized} & \textbf{Base} & \textbf{Self-Specialized} \\ 
\midrule\midrule
\multirow{1}{*}{\shortstack{SUM}} 
 & EDT-Summarization & \ \ 6.40 & \textbf{21.90} & 13.97 & \textbf{24.00} & 13.87 & \textbf{23.56} \\
\cmidrule(lr){1-8}
\multirow{2}{*}{\shortstack{QA}} 
 & InsuranceQA & \ \ 3.03 & \textbf{19.87} & \ \ 6.55 & \textbf{23.79} & \ \ 9.96 & \textbf{24.36} \\
 & ConvFinQA & \textbf{15.74} & \ \ 5.25 & \textbf{21.69} & 11.84 & \textbf{28.77} & 20.88 \\
\cmidrule(lr){1-8}
\multirow{2}{*}{\shortstack{NER}} 
 & Fin3 & \ \ 9.94 & \textbf{23.93} & \ \ 7.53 & \textbf{26.95} & \ \ 6.80 & \textbf{43.87} \\
 & FiNER\_139 & 10.24 & \textbf{14.84} & \textbf{36.78} & 25.81 & \textbf{44.34} & 35.63 \\
\cmidrule(lr){1-8}
\multirow{1}{*}{\shortstack{RE}} 
 & KPI-EDGER & 11.22 & \textbf{31.02} & 43.28 & \textbf{53.56} & 49.46 & \textbf{63.90} \\
\cmidrule(lr){1-8}
\multirow{3}{*}{\shortstack{SA}} 
 & EarningsCall & 46.80 & \textbf{48.80} & \textbf{50.80} & 48.00 & \textbf{49.03} & 47.74 \\
 & Financial\_Phrasebank & 23.60 & \textbf{73.20} & \ \ 9.40 & \textbf{47.60} & 29.20 & \textbf{68.80} \\
 & FIQA-SA & 44.44 & \textbf{56.84} & 58.55 & \textbf{61.54} & 61.54 & \textbf{70.09} \\
\cmidrule(lr){1-8}
\multirow{1}{*}{\shortstack{CLS}} 
 & Gold Commodity News & 21.95 & \textbf{43.03} & \textbf{61.93} & 55.08 & 38.42 & \textbf{61.20} \\
\midrule
\multicolumn{2}{c}{Average\qquad\qquad\qquad} & 19.34 & \textbf{33.87} & 31.05 & \textbf{37.82} & 33.14 & \textbf{46.00} \\
\bottomrule
\end{tabular}
\end{adjustbox}
\caption{Comparative results (\textsc{\textit{$F_1$}-Score}) of the base LM and self-specialized one on biomedical (top) and financial (bottom) domains. The base model is MPT-30B for biomedicine and LLaMA-2 7B for finance. Self-specialized ones have the same parameters as the counterpart base ones. \textit{k} indicates the number of demonstrations in a prompt.}
\label{table:main_results}
\end{table*}

\section{Experimental Settings}
\label{subsec:setups}
\paragraph{Datasets.} For our primary evaluation, we employ various biomedical NLP datasets, most of which are curated in \textsc{BigBio} \citep{NEURIPS2022_a583d219}. A total of 10 different datasets are adopted to encompass a wide range of NLP tasks: Question Answering (QA), Named Entity Recognition (NER), Relation Extraction (RE), Sentiment Analysis (SA), and Document Classification (DC). 
Following a prior work \citep{parmar2022boxbart}, all datasets are transformed into instructional data. 
Additionally, we validate our method in the financial domain to showcase its generalizability. We adopt a total of 10 diverse datasets, covering numerous NLP tasks: Summarization (SUM), QA, NER, RE, SA, and Classification (CLS), 
detailed in Appendix \ref{appendix:datasets}.

\paragraph{Models.} We employ MPT-30B \citep{MosaicML2023Introducing} as a base model for main experiments. 
For the retriever, we use simple yet effective BM25 \citep{Robertson1994OkapiAT}, assuming human-labeled data is not sufficient.
For benchmarking of general-purpose aligned models, we evaluate Alpaca-65B \citep{alpaca} and Dromedary-65B \citep{sun2023principledriven} that are both based on LLaMA \citep{touvron2023llama}. In addition to MPT-30B, we adopt LLaMA-2 7B \citep{touvron2023llama2} and Falcon-40B \citep{falcon40b} to further validate the general applicability of self-specialization with different scales and base models. We additionally evaluate existing domain-specific models \citep{wu2023pmcllama}: MedLLaMA and PMC-LLaMA (Details are in Section \ref{subsec:analyses}).

\paragraph{Metrics.} In our study, all tasks are approached as a unified text generation problem, aiming to assess the capabilities of generative models. In alignment with an established convention \citep{parmar2022boxbart}, we adopt \textsc{$F_1$-Score} as our main evaluation metric, given an early observation that \textsc{Rouge-L} \citep{lin-2004-rouge}, as shown in Table \ref{table:main_results_rouge} in Appendix, exhibits a strong correlation with \textsc{$F_1$-Score}.

\paragraph{Implementation Details.} For biomedical seeds, we use data sampled from BoX \citep{parmar2022boxbart}, encompassing 32 tasks, up to 5 instances for each dataset, resulting in a compact yet representative 80 seed samples in total, which are also used as demonstrations at inference. For optional external corpus, we leverage PubMed
preprocessed in \citep{phan2021scifive}, which contains $\approx$30M abstracts. In the financial domain, based on our finding from biomedical experiments that showed surprising effectiveness of self-specialization relying on internal knowledge of LLMs without the external corpus, we opt not to employ an optional retrieval component to further validate the self-sufficiency of LLMs. We leverage a total of 90 seeds sampled from the 10 train sets in our corresponding benchmark datasets. 
We use a total of 5K synthetic data generated through our self-specialization for all experiments, unless otherwise specified.
Being equipped with QLoRA \citep{dettmers2023qlora} and 4-bit quantization, the model is trained using a simple Alpaca-style template \citep{alpaca} on a single A100, taking only a few hours for 3 epochs, resulting in a light-weight specialization module.

\begin{figure*}[th!]
    \centering
    \includegraphics[width=1\textwidth]
    {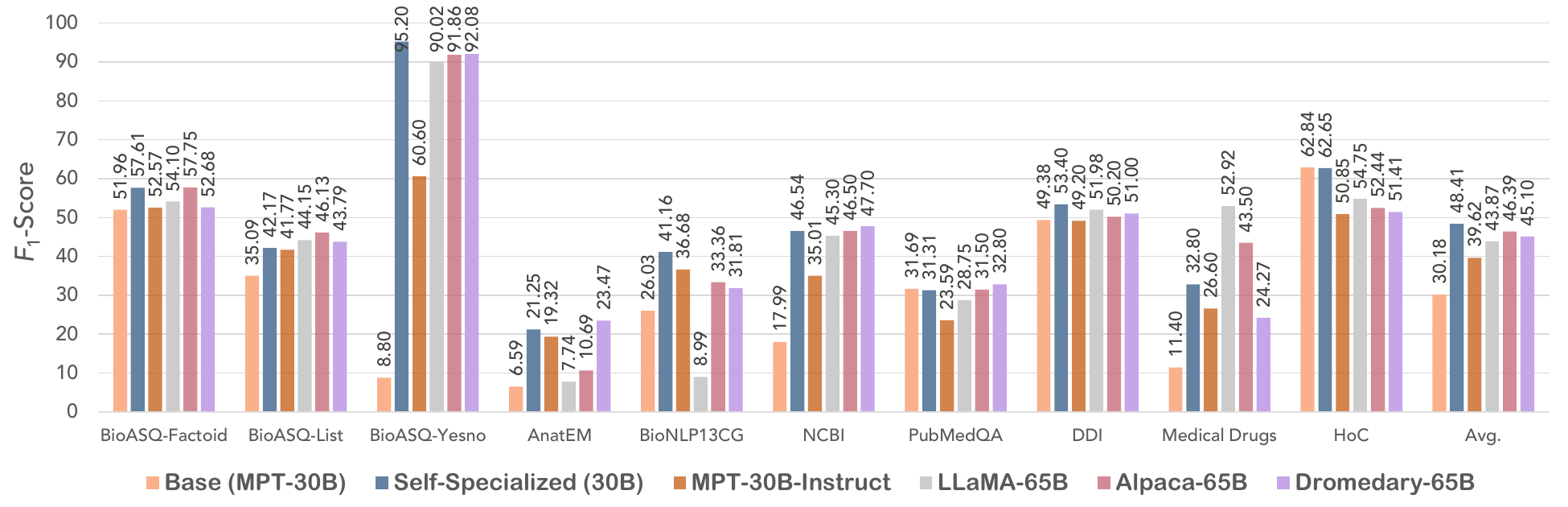}
    \caption{Comparing (with \textsc{\textit{$F_1$}-Score}, 5-shot) our self-specialized MPT-30B model to 65B models in biomedicine.}
    \label{fig:benchmark}
\end{figure*}

\section{Results and Analyses}
Here, we provide a set of experimental results and analyses
to address relevant research questions.
\subsection{Comparison with Baselines}
\label{subsec:main_result}
\paragraph{How effective is the self-specialization of base models?} In Table \ref{table:main_results}, we present the comparative results of our self-specialized model against its base counterpart across 10 distinct biomedical NLP and 10 financial NLP datasets.
The evaluation is conducted with varying numbers of in-context demonstrations, \textit{k}.

Our findings reveal that the self-specialized model exhibits remarkable progress in the majority of tasks across all configurations in both domains, yielding a substantial (up to 18 points) improvement in average scores. Specifically, the average scores ($F_1$) in biomedicine rise 
from 30.18 to 48.41 in a 5-shot setting.\footnote{\hypertarget{bioasq-footnote}Even excluding BioASQ-Yesno as an outlier due to the base model's low performance, self-specialization still shows significant gain over the base model: 32.55 to 43.21 (5-shot). Appendix \ref{subsec:evaluation_desings} includes the detailed discussion.}
In finance, the improvements are 14.53 (0-shot), 6.77 (1-shot), and 12.86 (5-shot), respectively. 
These advancements in both domains underscore the self-specialization's generalizability in addressing a wide array of tasks across different specialized domains.


\paragraph{Imact on ICL capability.}
A potential concern on self-specialization tuning is its impact on the base LLM's in-context learning capabilities, as we did not tune the model with demonstrations. Comparing the capabilities before and after self-specialization, the improvement after adding demonstrations (from 0 to 5) of our self-specialized model on biomedicine in Table \ref{table:main_results} is 36.63 to 48.41 ($\Delta$=11.78), while that of the base model is 25.15 to 30.18 ($\Delta$=5.03), indicating even better ICL capability with in-domain knowledge acquisition.

\paragraph{Performance drop on some tasks.}
Our analysis does identify a few instances where performance drops as shown in Table \ref{table:main_results}. This indicates room for further refinement, especially for tasks like ConvFinQA that require a set of specific capabilities beyond mere domain knowledge. We evidenced that a minor proportion ($\leq$ 2\%) of generated data partially resembles ConvFinQA, due to our generation's nature involving creative brainstorming for diversity. The specific demands of ConvFinQA, including numerical reasoning, structured tables, and conversations extend beyond basic domain knowledge and were insufficiently covered within our dataset. This gap likely contributes to the observed performance trade-offs.

\begin{figure}[!t]
    \centering
    \includegraphics[width=0.48\textwidth]
    {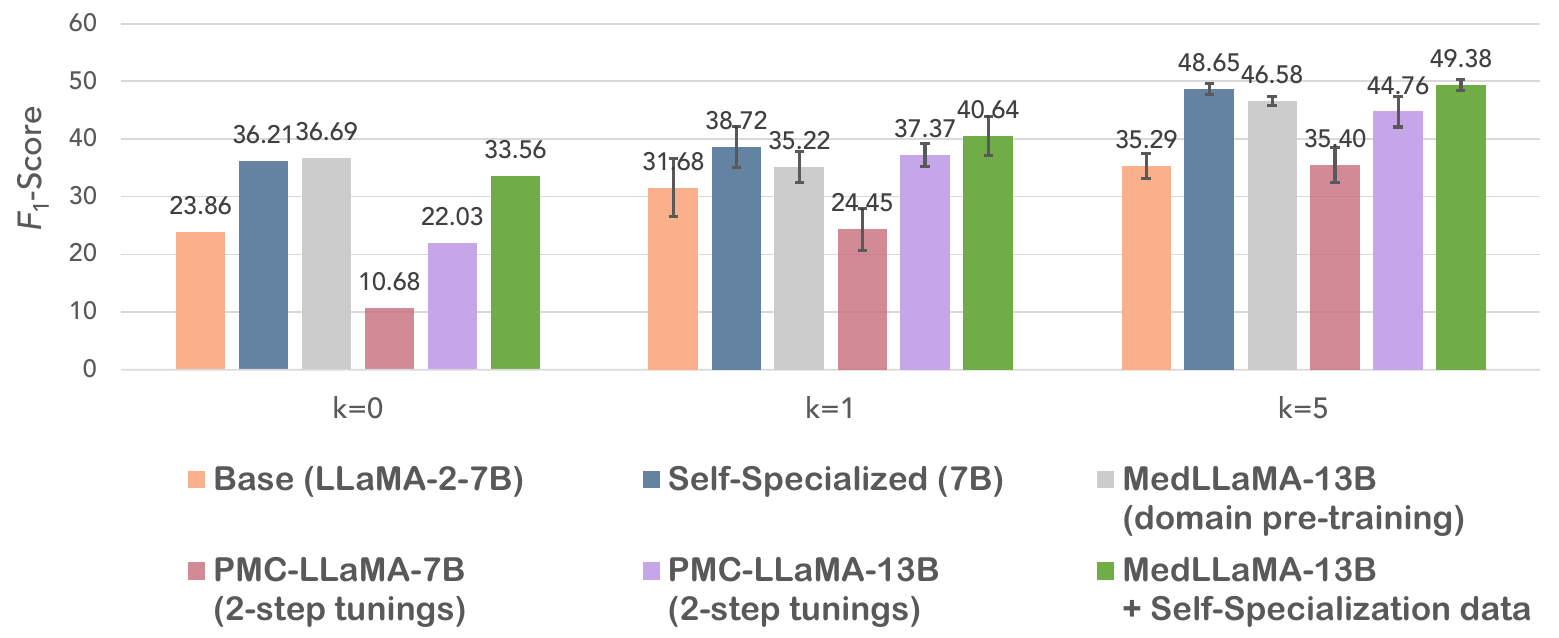}
    \caption{Results in biomedicine using LLaMA-2 7B as a base model, and comparisons with other baselines including the one pre-trained on a huge domain-specific corpus. Scores are averaged over 10 datasets, and when in-context examples are involved, we use 5 different sets of demonstrations to report macro-averaged results and variances (SD) with error bars.}
    \label{fig:7b_bio}
\end{figure}

However, we re-emphasize that there are significantly bigger gains in many of the cases (e.g., 45 out of 60 experiments across datasets and \textit{k}), outweighing the regression overall.
Acknowledging the inherent variability of in-context learning \citep{min-etal-2022-metaicl}, we present the variances with 5 different sets of demonstrations in Figure \ref{fig:7b_bio} based on LLaMA-2-7B in biomedicine, showing significant average improvements of 8.25 ($p=0.003, k=1$) and of 14.42 ($p\leq0.001, k=5$).

\paragraph{How does self-specialization compare against larger/generally aligned baselines?} In Figure \ref{fig:benchmark}, we compare our self-specialized MPT-30B model with 65B models, including LLaMA-65B, and its general instruction aligned variants (e.g., Alpaca based on Self-Instruct) in the biomedical domain. 
Interestingly, the results reveal that our model, without extensive data, surpasses all baselines, including 65B models, despite its $\approx$2.2x smaller size. This not only highlights the lower expert domain performance trade-offs of the ``generalist'' models in terms of encoding vast general knowledge into a finite set of parameters, but also underscores the effectiveness of our parameter-efficient approach to model specialization. We also show that our self-specialized model outperforms the supervised general-purpose model, MPT-30B-instruct in all tasks, which highlights the benefits of in-domain instruction data. 
Moreover, as a reference point, we present a comparison with a fully-supervised SOTA model that is fine-tuned on the biomedical datasets in Table \ref{table:SOTA}, contextualizing our progress to better understand practical utility, discussed in Appendix \ref{subsec:SOTA_performances}.
Notably, the data efficiency of our simple self-specialization is further reinforced by the fact that the model is trained using only 5K\footnote{52K for Alpaca and 360K for Dromedary.} instruction data self-produced with minimal (only 80) seeds.\footnote{175 for Alpaca and 195 for Dromedary.} This training process, facilitated by the incorporation of QLoRA, adding only 0.28\% trainable parameters to an otherwise frozen model, only takes a few hours on a single GPU (A100 80GB).

\begin{table}[t!]
\begin{center}
\begin{adjustbox}{width=0.4\textwidth}
\begin{tabular}{lcc}
\toprule
  Model & \textsc{$F_1$-Score}  & \textsc{Rouge-L} \\
  \midrule\midrule
w/ Top-5 Docs & \textbf{34.57} & \textbf{32.88} \\ 
w/ Top-1 Docs & 29.65 & 27.90  \\
w/o Retrieval & 33.72 & 32.14 \\
\cmidrule(lr){1-3}
Base MPT-30B & 25.15 & 23.75 \\
\bottomrule
\end{tabular}
\end{adjustbox}
\end{center}
\caption{Ablation of self-specialization with retrieval from unlabeled domain-specific documents. 
Zero-shot average performance over 10 biomedical tasks.
}
\label{table:ablation_retrieval}
\end{table}


\subsection{Ablations \& Analyses}
\label{subsec:analyses}
\paragraph{Effect of external knowledge.} We investigate the influence of incorporating a domain-specific corpus like PubMed in the response generation phase. 
Table \ref{table:ablation_retrieval} shows optimal results with the top-5 documents, while using just the top-1 document decreases performance, likely due to noise from an imperfect retrieval process, aligned with findings from previous work \citep{yoran2023making} that adding irrelevant (i.e., random) context dramatically decreases performances. Conversely, employing the top-5 documents with probability marginalization (eq. \ref{eq:margin}) seems to mitigate this issue, enabling the model to exploit informative knowledge. 
Interestingly, we observe that self-specialization demonstrates strong performance even without retrieval, 
suggesting domain-relevant knowledge is intermingled with other information acquired during pre-training, which self-specialization uncovers to better utilize.
Given this, the added complexity of retrieval mechanisms, though potentially advantageous, emerges as optional within our framework.

\begin{table}[t!]
\begin{center}
\begin{adjustbox}{width=0.4\textwidth}
\begin{tabular}{lcc}
\toprule
  Model & \textsc{$F_1$-Score}  & \textsc{Rouge-L} \\
  \midrule\midrule
    w/ Iterative Process & \textbf{36.63} & \textbf{34.79} \\
    Self-Specialization & 34.57 & 32.88 \\
    \cmidrule(lr){1-3}
    Base MPT-30B & 25.15 & 23.75 \\
\bottomrule
\end{tabular}
\end{adjustbox}
\end{center}
\caption{Ablation of iterative self-specialization. 
Zero-shot average performance over 10 biomedical tasks.
}
\label{table:ablation_iter}
\end{table}

\paragraph{Effect of iterative self-specialization.} In Section \ref{subsec:iter}, we discussed the potential of employing an iterative process by leveraging the self-specialized model instead of the base model throughout the generation process. 
Table \ref{table:ablation_iter} shows the ablation study, where each iteration involved generating 5K samples, and final results were obtained using 5K samples from the last iteration for a fair comparison. We observe that the iterative process leads to further performance enhancements, compared to the one w/o iteration. In our preliminary tests, we rarely find meaningful improvements with the subsequent iteration, which we leave for future work to refine.

\paragraph{Self-specialization vs. domain pre-training.}
We compare our model based on LLaMA-2-7B with existing baselines \citep{wu2023pmcllama}: MedLLaMA-13B and PMC-LLaMA-7B/-13B. The former is an LLaMA variant further pre-trained on a large domain-specific corpus (i.e., medicine), and the latter is further instruction-tuned using annotated/synthetic datasets, including medical QA, rationale for reasoning, and conversational dialogues. Notably, we find that our self-specialized 7B model is on par with or better than MedLLaMA-13B ($p=0.006, k=5$) and PMC-LLaMA-13B ($p=0.01, k=5$) despite their larger parameters and extensive domain-specific tuning. 
Additionally, using our 7B-generated data to specialize MedLLaMA indicates that self-specialization can enhance domain-specific pre-training ($p=0.001, k=5$), suggesting complementarity.


\begin{figure}[!t]
    \centering
    \includegraphics[width=0.45\textwidth]
    {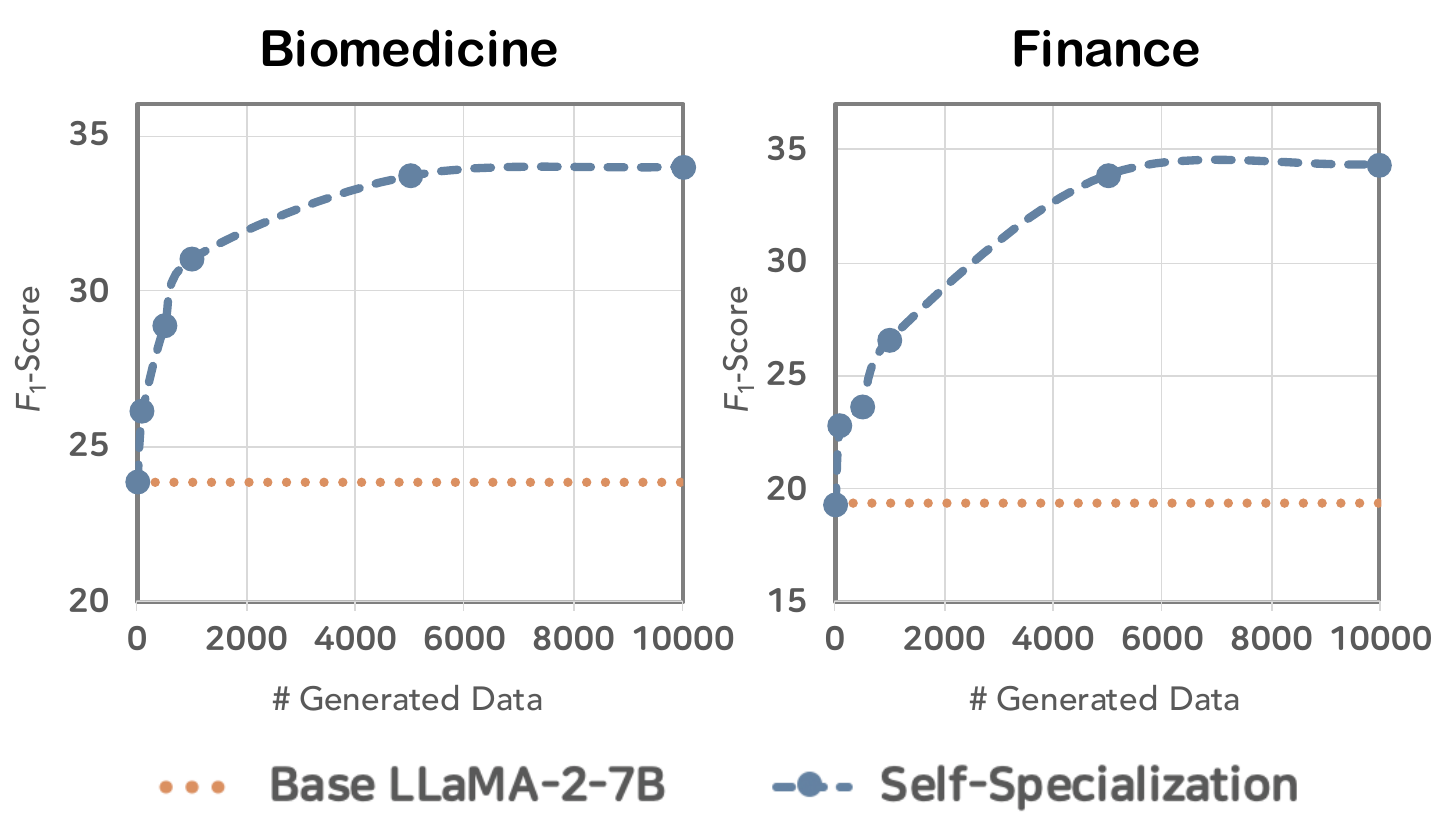}
    \caption{Analysis with the varied number of self-generated data for specialization. 0-shot averaged results with \# generated data \texttt{\small{= \{0, 100, 500, 1000, 5000, 10000\}}} are shown.}
    \label{fig:data_analysis}
\end{figure}

\begin{figure*}[!th]
    \centering
    \includegraphics[width=0.53\textwidth]
    {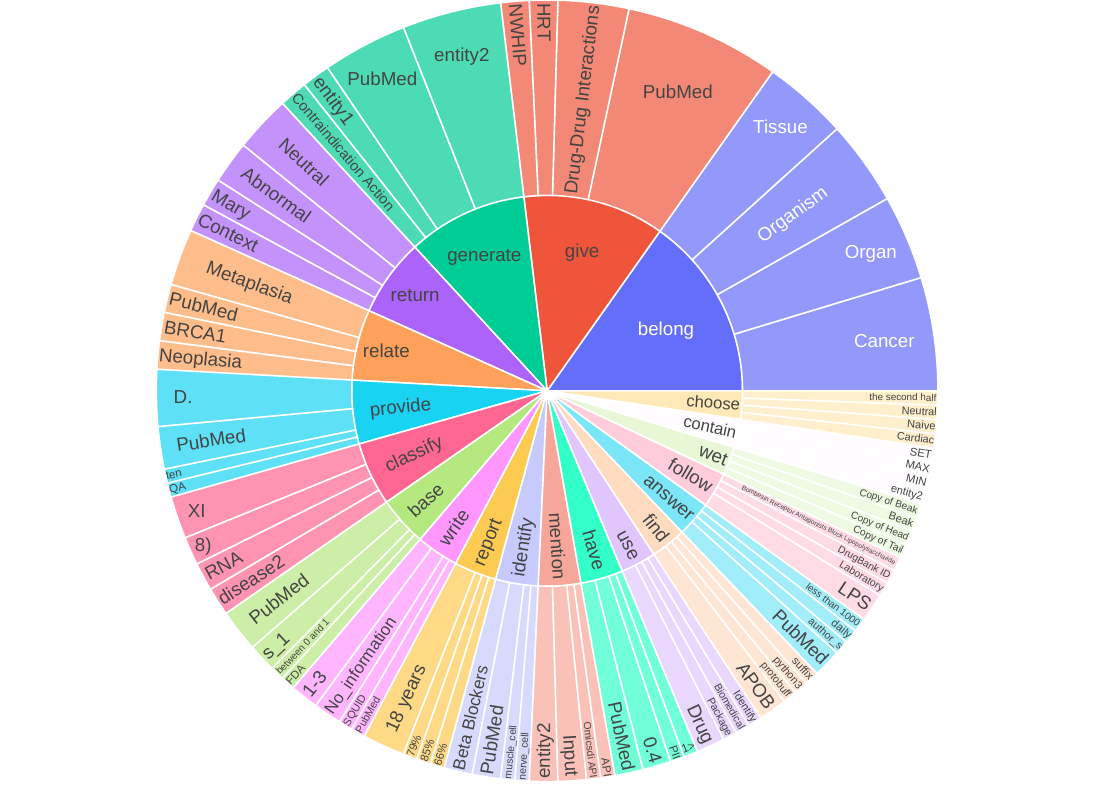}
    \includegraphics[width=0.4\textwidth]
    {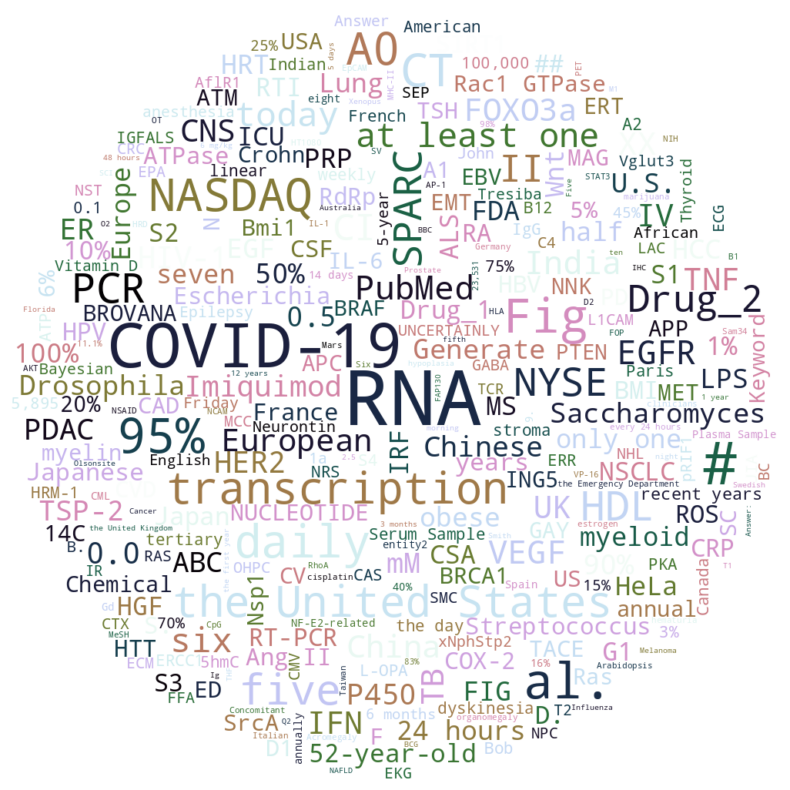}
    \caption{Statistics for generated data through self-specialization. On the left, the inner circle illustrates prevalent verbs in the instructions, with the outer ring revealing associated entities. Conversely, the right side showcases the input context, highlighting the incorporation of diverse biomedical keywords. Best viewed in zoom and color.}
    \label{fig:stats}
\end{figure*}

\paragraph{Impact of the number of self-generated data.} 
In Figure \ref{fig:data_analysis}, we analyze the impact of the number of self-specialization data within biomedical and financial domains. Starting from zero, a sharp increase in $F_1$ score is observed as we introduce the first 100 instances which largely consist of seed instructions, underlining the significant impact of seeds not only as in-context demonstrations but also as training data. The performance continues to rise steadily with additional data, plateauing around 5K instances, supporting our decision on the use of 5K data. 
Self-specialization's success with relatively small self-generated data highlights its data efficiency and practicality.

\paragraph{How is the quality of synthetic self-specialization data?} 
In Figure \ref{fig:stats}, we showcase a qualitative visualization that analyzes the synthetic data generated through self-specialization, confirming that self-specialization produces domain-focused data. To quantitatively assess the quality, Figure \ref{fig:falcon} in Appendix compares our model against a model trained on labeled data, which shows a narrow performance gap, implying the quality of generated data.
Additionally, some examples are provided in Table \ref{table:examples1} \& \ref{table:examples2} in Appendix, offering insights into the quality of the self-generated specialization data.

\section{Related Work}
The goal of instruction-tuning and alignment of large language models (LLMs) is to achieve cross-task generalization or to align with human preferences. This can be accomplished by either training LLMs directly with human-labeled data \citep{ouyang2022training, wei2022finetuned, naturalinstructions, supernaturalinstructions} or data generated by larger models (i.e., distillation) \citep{alpaca, vicuna2023}. Recent studies have shown that LLMs are self-instructors. \citet{wang2022self} showed that with in-context prompts, GPT-3 \citep{NEURIPS2020_1457c0d6} can generate high-quality instruction-responses pairs for its own alignment. \citet{sun2023principledriven} further suggests that using principles can minimize human supervision while covering a broad spectrum of scenarios with the open-source model, LLaMA-65B \citep{touvron2023llama}. 
While enhancing general alignment, according to our presented evidence, these approaches are unlikely to induce specialization in expert domains, leaving different domain expertise in superposition inside the model. 
To the best of our knowledge, we are the first to show the potential for expert domain specialization through self-alignment, effectively ``uncovering'' a domain expert out of the model in a parameter- and data-efficient manner.

Recent studies highlight the benefits of employing instructions in different adaptation scenarios \citep{parmar2022boxbart}. \textsc{InstructOR} \citep{su2022one} illustrated the adaptability of instruction-based text embeddings to various tasks and domains, while \textsc{InstrucTE} \citep{bai2023schemadriven} demonstrated that incorporating instructions with a schema can yield robust results for table extraction across diverse domains. However, these require the use of costly human labels or extensively tuned large models (e.g., 175B). 
Self-training has also been explored for different adaptation scenarios. For domain knowledge adapation, \citet{shakeri2020end} and \citet{luo2022cooperative} proposed constructing synthetic data by generating in-domain question-answering data, but these data generators are trained with more than 80k human curated QA pairs and do not involve instructional ones that have the potential for cross-task generalization. Instruction-tuning has been shown to adapt pre-trained LLMs to different modalities, including vision \citep{liu2023visual}, audio \citep{gong2023listen}, and programs \citep{roziere2023code}, and enables the use of APIs \citep{schick2023toolformer} and search engines \citep{luo2023sail}.
Unlike these works, our work focuses on uncovering target domain expertise latent within LLMs while promoting cross-task generalization with minimal supervision.

\section{Conclusion}
Our exploration into self-specialization
aimed to elucidate the latent expertise within large language models (LLMs) with limited human supervision. This scheme
demonstrated promising results in specialized domains. The self-specialized model exhibited remarkable performance, outperforming its base model, MPT-30B, and even surpassing larger generally aligned models (65B). This illuminates the intrinsic challenges of encoding vast general knowledge into limited parameters and underscores the efficiency of self-specialization. Remarkably, the model’s efficient training, marked by minimal data usage and the integration of QLoRA \citep{dettmers2023qlora}, adds another layer to its practicality in terms of parameter and data efficiency. These findings signify a promising pathway for leveraging inherent expertise in LLMs and offering a large variety of exciting opportunities for future work.

\section*{Limitations}
While our study provides encouraging insights into the capabilities of self-specialization, this is an initial step in opening up new opportunities. 
We acknowledge the need for further exploration and note some limitations and considerations.

\paragraph{Sensitivity of in-context learning.}
In Table \ref{table:main_results}, we observed that performances sometimes dropped with more in-context learning demonstrations. While recognizing, the performance fluctuation is not an issue stemming from our self-specialization tuning, as it happens for the base LLM as well as GPT-3 (Appendix \ref{subsec:sensitivity_of_prompting})
, demonstrating an inherent challenge in in-context learning. This phenomenon is not unique to our self-specialization approach, but a broader challenge in the field.

\paragraph{Training and generation strategies.}
We avoided using demonstrations during training \citep{min-etal-2022-metaicl} to maintain flexibility in the number of examples available during inference.
We aimed to ensure that zero-shot performance remains unaffected by tuning to rely on demonstrations.

Unlike previous work \citep{wang-etal-2023-self-instruct} that generates instructions first and then inputs/responses together, our approach simultaneously generates instructions and inputs, followed by responses. This strategy, inspired by a more recent work \citep{sun2023principledriven}, enables the use of inputs as queries for retrieval prior to response generation.
Despite the specific reasons outlined above, we recognize the potential of the alternative strategies as avenues for future exploration, which can be orthogonal to our current approach.


\paragraph{Filtering.}
In our method, we opted not to implement an automatic filtering process for the generated data. 
In a preliminary study to assess feasibility, we attempted to filter out low-quality data manually, however we did not observe a noticeable improvement. We hypothesized that incorporating this seemingly unuseful data may even enhance the model's robustness by preventing overfitting to those generated data. Despite this, we acknowledge the importance of further investigating filtering techniques for potential improvements.



\paragraph{Potential data contamination and bias propagation.}
Being cautious with potential data contamination from base language models during self-specialization, we conducted stringent measures following practices in GPT-3 \citep{NEURIPS2020_1457c0d6} and PaLM \cite{chowdhery2022palm}. We adopted n-gram overlap analysis (with n=8 and a threshold of 70\%) to scrutinize similarities between our generated data and all the test sets, revealing no significant overlaps. Moreover, a detailed manual inspection of 200 random instances corroborated this finding. 
When concerned about retrieval sources, one can apply the n-gram overlap filtering, though our sources are PubMed abstracts without explicit labels, which inherently ensures little risk of data overlap.
Meanwhile, we acknowledge the inherent risk of propagating biases from the pre-trained data. 


\section*{Acknowledgments}
This research is supported in part by the NSF (IIS-2052498), ODNI and IARPA via the HIATUS program (contract 2022-22072200004). The views and conclusions contained herein are those of the authors and should not be interpreted as necessarily representing the official policies, either expressed or implied, of NSF, ODNI, IARPA, or the U.S. Government. The U.S. Government is authorized to reproduce and distribute reprints for governmental purposes notwithstanding any copyright annotation therein.

\bibliography{anthology,custom}

\clearpage
\appendix
\section{Explanations of Evaluation Datasets}
\label{appendix:datasets}
Below are brief descriptions for each dataset in biomedical and financial domains. All datasets are in English.
\subsection{Biomedicine}
\paragraph{BioASQ-8b \citep{bioasq}.} This is a biomedical QA dataset that necessitates models to produce answers from given questions and corresponding contexts within the biomedical domain. There are three distinct subsets that can be divided according to question types: Factoid, List, and Yesno. This dataset is publicly available
upon a data use agreement. The data are originally intended to be used as training and development data, and we use the small part of the training set as seeds (i.e., 5 seeds), and the test set for evaluation (500 for each question type). CC BY 2.5.

\paragraph{PubMedQA-Long \citep{jin2019pubmedqa}.} PubMedQA is another biomedical QA dataset featuring research questions along with their corresponding abstracts and answers sourced from PubMed\footnote{https://www.ncbi.nlm.nih.gov/pubmed}. To diversify the task types, we focus on a long-form answer (i.e., conclusion). We use 5 labeled data for seeds and 500 for evaluation. MIT license.

\paragraph{AnatEM \citep{10.1093/bioinformatics/btt580}.} This is a Named Entity Recognition (NER) task for anatomical entities in biomedical texts. Models are tasked with identifying all anatomy-named entities and their corresponding types from given a small paragraph. Non-commercial purposes only. 404 test data are used for evaluation and 5 training instances are used for seeds. CC BY-SA 3.0.

\paragraph{BioNLP13CG \citep{pyysalo-etal-2013-overview}.} The Cancer Genetics (CG) is an information extraction task targeting the recognition of events in text, encompassing multiple levels of biological organization, from molecular to whole organisms. 5 training data are used for seeds, and the number of evaluation data is 200. CC BY-SA 3.0.

\paragraph{NCBI \citep{ncbi}.} The NCBI disease corpus, derived from the National Center for Biotechnology Information, focuses on disease name recognition. According to the annotation guideline of this dataset, organism names such as humans, and also gender are excluded for annotation. We use 5 training instances for seeds, and 100 for evaluation. The data is freely available to the public for use. CC0 1.0 license.

\paragraph{DDI \citep{ddi}.} The Drug-Drug Interaction (DDI) dataset is tailored for identifying interactions between different drugs in biomedical texts. Following \citet{parmar2022boxbart}, this work considers only binary Relation Extraction (RE), determining whether there is an effect of given two drugs. The data cannot be used for any commercial purposes. We use 5 data for seeds, and 500 for evaluation. CC BY-NC 4.0.

\paragraph{Medical Drugs \citep{medicaldrugs}.} This is a Sentiment Analysis (SA) dataset that is required to predict the sentiment of individuals towards medical drugs. Specifically, given a text and a drug, a model determines the effect of the drug as ``positive", ``negative", or ``neutral". 5 training instances are used for the seed construction, and 500 test set for evaluation. The license is unknown.

\paragraph{HoC \citep{10.1093/bioinformatics/btv585}.} The Hallmarks of Cancer (HoC) dataset is curated for classifying (zero to many) biomedical texts related to cancer into categories representing different hallmarks of cancer. In particular, these hallmarks include ``sustaining proliferative signaling", ``resisting cell death", ``genomic instability and mutation", ``activating invasion and metastasis", ``tumor promoting inflammation", ``evading growth suppressors", ``inducing angiogenesis", ``enabling replicative immortality", ``avoiding immune destruction" and ``cellular energetics". The number of evaluation data is 200 and 5 training data are used for seed demonstrations. GPL-3.0 license.

\subsection{Finance}
\paragraph{EDT-Summarization \citep{zhou-etal-2021-trade}.} This dataset challenges models to perform abstractive summarization on financial news articles, condensing detailed information into succinct summaries. 8 training instances are used for seeds, and 500 instances for evaluation. This data is publicly available.

\paragraph{InsuranceQA \citep{InsuranceQA}.} This is an open-book question-answering task about insurance, demanding models to extract and provide specific insurance-related information. Seed demonstrations include 8 training data and the number of evaluation instances is 500. This dataset is provided as is and for research purposes only. 

\paragraph{ConvFinQA \citep{chen2022convfinqa}.} This is a dataset for conversational question-answering over financial report tables, testing a model's ability to reason and respond within a conversational context. We use 8 training data for the seed construction, and evaluation uses 500 test instances. MIT license.

\paragraph{Fin3 \citep{salinas-alvarado-etal-2015-domain}.} This is a financial NER dataset based on financial agreements to aid credit risk assessments. 8 training data are used for seeds and 100 test data for evaluation. CC-BY 3.0.

\paragraph{FiNER\_139 \citep{loukas-etal-2022-finer}.} This NER task focuses on financial texts, where models identify and classify financial-related entities like numbers. This dataset includes a much larger label set of 139 entity types. Seed data encompass 8 training instances and the number of test data is 500. MIT license.

\paragraph{KPI-EDGAR \citep{deusser2022kpiedgar}.} Models are tasked with extracting key performance indicators (KPIs) from financial documents. Categories for KPIs include current and previous year values, annual changes, subordinate and descriptive attributes, co-references, and false-positive. We use 212 test instances for evaluation and 8 training instances for seed demonstrations. MIT license.

\paragraph{EarningsCall \citep{202102.0424}.} This is a binary sentiment analysis task where models evaluate sentiments from stock values and transcripts of earnings calls, reflecting the financial sentiments expressed. 8 training instances are used for seeds, and 500 test set for evaluation. CC0 1.0 license.

\paragraph{Financial\_Phrasebank \citep{Malo2014GoodDO}.} This dataset involves (3-way) sentiment analysis of financial news headlines, assessing the underlying sentiment conveyed by the language used. Commercial uses of this data may be allowed upon contacting the authors. 8 training data and 500 test data used for seeds, and evaluation, respectively. CC BY-NC-SA 3.0.

\paragraph{FIQA-SA \citep{Maia2018WWW18OC}.} It consists of aspect-based sentiment analysis tasks within financial texts, requiring models to discern sentiment regarding specific aspects mentioned. The number of evaluation data is 234 and seed demonstrations include 8 training instances.

\paragraph{Gold Commodity News \citep{CommodityNews}.} This dataset involves classifying financial news headlines about gold commodities into categories such as market movement direction or type of financial news (e.g., direction up, down, pastprice, futurenews, etc). The seed data includes 9 binary-class version and also 9 multi-class version of training set, and evaluation uses 500 multi-class version of test data. The license of this data indicates data files © original authors.

\begin{table*}[t!]
\centering
\begin{adjustbox}{width=1.0\textwidth}
\begin{tabular}{llcccccc}
\toprule
\multicolumn{2}{c}{\textsc{Biomedicine}} &  
\multicolumn{2}{c}{Worst} & \multicolumn{2}{c}{Average} & \multicolumn{2}{c}{Best} \\ 
\cmidrule(lr){3-4}\cmidrule(lr){5-6}\cmidrule(lr){7-8}
 \textbf{Task} & \textbf{Dataset} & \textbf{Base} & \textbf{Self-Specialized} & \textbf{Base} & \textbf{Self-Specialized} & \textbf{Base} & \textbf{Self-Specialized} \\ 
\midrule\midrule
\multirow{4}{*}{\shortstack{QA}} 
 & BioASQ-Factoid & 30.90 & \textbf{37.35} & 43.47 & \textbf{50.00} & 51.96 & \textbf{57.61} \\
 & BioASQ-List & 35.09 & \textbf{42.17} & 42.91 & \textbf{44.57} & \textbf{47.57} & 46.99 \\
 & BioASQ-Yesno & \ \ 8.80 & \textbf{85.27} & 13.60 & \textbf{91.49} & 21.20 & \textbf{95.20} \\
 & PubMedQA & 11.98 & \textbf{24.16} & 24.19 & \textbf{26.78} & \textbf{31.69} & 31.31 \\
\cmidrule(lr){1-8}
\multirow{3}{*}{\shortstack{NER}} 
 & AnatEM & \ \ 6.59 & \textbf{11.99} & \ \ 7.93 & \textbf{16.33} & \ \ 9.63 & \textbf{21.25} \\
 & BioNLP13CG & 21.76 & \textbf{24.93} & 24.19 & \textbf{32.63} & 26.03 & \textbf{41.16} \\
 & NCBI & \textbf{17.99} & 14.35 & 21.44 & \textbf{34.67} & 27.88 & \textbf{46.54} \\
\cmidrule(lr){1-8}
\multirow{1}{*}{\shortstack{RE}} 
 & DDI & \textbf{49.20} & 49.40 & 49.86 & \textbf{51.47} & 51.00 & \textbf{53.40} \\
\cmidrule(lr){1-8}
\multirow{1}{*}{\shortstack{SA}} 
 & Medical Drugs & 11.40 & \textbf{32.80} & 19.27 & \textbf{51.07} & 35.00 & \textbf{65.80} \\
\cmidrule(lr){1-8}
\multirow{1}{*}{\shortstack{DC}} 
 & HoC & \ \ 2.44 & \textbf{\ \ 6.01} & \textbf{26.40} & 25.42 & \textbf{62.84} & 62.65 \\
\midrule
\multicolumn{2}{c}{Average\qquad\qquad\qquad} & 19.62 & \textbf{32.84} & 27.33 & \textbf{42.44} & 36.48 & \textbf{52.19} \\
\bottomrule
\end{tabular}
\end{adjustbox}
\smallskip

\begin{adjustbox}{width=1.0\textwidth}
\begin{tabular}{llcccccc}
\toprule
\multicolumn{2}{c}{\textsc{Finance}\qquad\qquad} &  
\multicolumn{2}{c}{Worst} & \multicolumn{2}{c}{Average} & \multicolumn{2}{c}{Best} \\ \cmidrule(lr){3-4}\cmidrule(lr){5-6}\cmidrule(lr){7-8}
 \textbf{Task} & \textbf{Dataset} & \textbf{Base} & \textbf{Self-Specialized} & \textbf{Base} & \textbf{Self-Specialized} & \textbf{Base} & \textbf{Self-Specialized} \\ 
\midrule\midrule
\multirow{1}{*}{\shortstack{SUM}} 
 & EDT-Summarization & \ \ 6.40 & \textbf{21.90} & 11.41 & \textbf{23.15} & 13.97 & \textbf{24.00} \\
\cmidrule(lr){1-8}
\multirow{2}{*}{\shortstack{QA}} 
 & InsuranceQA & \ \ 3.03 & \textbf{19.87} & \ \ 6.51 & \textbf{22.67} & \ \ 9.96 & \textbf{24.36} \\
 & ConvFinQA & \textbf{15.74} & \ \ 5.25 & \textbf{22.07} & 12.66 & \textbf{28.77} & 20.88 \\
\cmidrule(lr){1-8}
\multirow{2}{*}{\shortstack{NER}} 
 & Fin3 & \ \ 6.80 & \textbf{23.93} & \ \ 8.09 & \textbf{31.58} & \ \ 9.94 & \textbf{43.87} \\
 & FiNER\_139 & 10.24 & \textbf{14.84} & \textbf{30.45} & 25.43 & \textbf{44.34} & 35.63 \\
\cmidrule(lr){1-8}
\multirow{1}{*}{\shortstack{RE}} 
 & KPI-EDGER & 11.22 & \textbf{31.02} & 34.65 & \textbf{49.49} & 49.46 & \textbf{63.90} \\
\cmidrule(lr){1-8}
\multirow{3}{*}{\shortstack{SA}} 
 & EarningsCall & 46.80 & \textbf{47.74} & \textbf{48.88} & 48.18 & \textbf{50.08} & 48.80 \\
 & Financial\_Phrasebank & \ \ 9.4 & \textbf{47.60} & 20.73 & \textbf{63.20} & 29.20 & \textbf{73.20} \\
 & FIQA-SA & 44.44 & \textbf{56.84} & 54.84 & \textbf{62.82} & 61.54 & \textbf{70.09} \\
\cmidrule(lr){1-8}
\multirow{1}{*}{\shortstack{CLS}} 
 & Gold Commodity News & 21.95 & \textbf{43.03} & 40.77 & \textbf{53.10} & \textbf{61.93} & 61.20 \\
\midrule
\multicolumn{2}{c}{Average\qquad\qquad\qquad} & 17.60 & \textbf{31.20} & 27.84 & \textbf{39.23} & 35.99 & \textbf{46.59} \\
\bottomrule
\end{tabular}
\end{adjustbox}
\caption{Comparative results of the base LM and self-specialized one on a biomedical domain (top) and on a financial domain (bottom). The base model is MPT-30B for biomedicine and LLaMA-2 7B for finance. Self-specialized ones have the same parameters as the counterpart base model. Performances are reported using \textsc{\textit{$F_1$}-Score}. The results are presented using worst, average, and best across 0-, 1-, and 5-shot results for each dataset.
}
\label{table:main_results_best}
\end{table*}

\section{Details of Experiments}
In Table \ref{table:prompts}, we show the prompts used for our self-specialization. For instruction generation, we leverage the prompt designed in self-instruct \citet{wang2022self} with minimal change to make it suit to specialization. In particular, we ask a model for instructions about a targeted domain, and force it to generate input together with the instruction, unlike in \citet{wang2022self} that generates those separately. In addition, we avoid using the specific requirement in the prompt that asks to cover diverse topics, such as (quoting \citet{wang2022self}) ``daily routines, travel and tourism health and wellness, cooking and recipes, personal finance, environmental issues, history and historical events, literature and literary analysis, politics and current events, psychology and mental health, art and design, mathematics and problem-solving, physics and astronomy, biology and life sciences, chemistry and materials science, computer science and programming, engineering and technology, robotics and artificial intelligence, economics and business management, philosophy and ethics, and more". For response generation, we use a simple prompt to let a model answer with a target domain in mind. 
Both prompts can be further enhanced and optimized for better self-specialization performance in future work.

Regarding our evaluations, we use prompt templates that were designed and used to optimize each Alpaca \citep{alpaca} and Dromedary \citep{sun2023principledriven}, but no specific template for base models, as they were not optimized for it during pre-training. Ours employs a simple Alpaca template for training and evaluation. We leverage publicly available delta weights that are supposed to be attached to LLaMA \citep{touvron2023llama} for Dromedary, and use the ones reproduced for Alpaca in our work.

We use three seed demonstrations in-context, which are randomly sampled from our initial seeds, and sampling with top-p being 0.98 and temperature being 1.0 during instruction generation. For response generation, we use no demonstrations in-context since there is a high chance that the generated instruction task and the sampled one do not match well. We believe further exploration of this aspect would be valuable in future work. For fine-tuning, we use a batch size of 32, a learning rate of 3e-4, and epochs of 3. Low-rank adaptation (LoRA) \citep{hu2022lora, dettmers2023qlora} is applied to all modules and all layers with a rank of 8, and an alpha of 16. While we report single-run results considering low-data settings where automatic hyperparameter tuning might be infeasible, we also report worst, average, and best across different k-shot configurations for each dataset to address the concern of sensitivity (Appendix \ref{subsec:sensitivity_of_prompting}) in Table \ref{table:main_results_best}.

\begin{table*}[t!]
\centering
\begin{adjustbox}{width=0.7\textwidth}
\begin{tabular}{llcc}
\toprule
 &  
& \multicolumn{2}{c}{\textsc{$F_1$-Score} / \textsc{Rouge-L}} \\ 
\cmidrule(lr){3-4}
 \textbf{Task} & \textbf{Dataset} & \textbf{Base} & \textbf{Self-Specialized} \\ 
\midrule\midrule
\multirow{4}{*}{QA} 
 & BioASQ-Factoid & 51.96 / 51.81 & \textbf{57.61} / \textbf{57.48} \\
 & BioASQ-List & 35.09 / 30.40 & \textbf{42.17} / \textbf{36.24} \\
 & BioASQ-Yesno & \ 8.80 / \ 8.80 & \textbf{95.20} / \textbf{95.20} \\
 & PubMedQA & \textbf{31.69} / 24.56 & 31.31 / \textbf{24.77} \\
\cmidrule(lr){1-4}
\multirow{3}{*}{NER} 
 & AnatEM & \ 6.59 / \ 6.07 & \textbf{21.25} / \textbf{19.24} \\
 & BioNLP13CG & 26.03 / 22.53 & \textbf{41.16} / \textbf{35.07} \\
 & NCBI & 17.99 / 16.60 & \textbf{46.54} / \textbf{41.55} \\
\cmidrule(lr){1-4}
RE 
 & DDI & 49.38 / 49.38 & \textbf{53.40} / \textbf{53.40} \\
\cmidrule(lr){1-4}
SA 
 & Medical Drugs & 11.40 / 11.40 & \textbf{32.80} / \textbf{32.80} \\
\cmidrule(lr){1-4}
DC 
 & HoC & \textbf{62.84} / \textbf{62.84} & 62.65 / 62.65 \\
\midrule
\multicolumn{2}{c}{Average} & 30.18 / 28.44 & \textbf{48.41} / \textbf{45.84} \\
\bottomrule
\end{tabular}
\end{adjustbox}
\caption{Comparative results (\textsc{\(F_1\)-Score} and \textsc{Rouge-L}) of the base LM and self-specialized one in the biomedical domain for \(k=5\). Scores are presented as \(F_1\) / ROUGE for each dataset. \textsc{Rouge-L} exhibits the same trend with \textsc{$F_1$-Score}.}
\label{table:main_results_rouge}
\end{table*}

\begin{table*}[t!]
\centering
\begin{adjustbox}{width=0.7\textwidth} 
\begin{tabular}{llccc}
\toprule
\textbf{Task} & \textbf{Dataset} & \textbf{Base} & \textbf{Self-Specialized} & \textbf{SOTA} \\ 
\midrule\midrule
\multirow{4}{*}{\shortstack{QA}} 
 & BioASQ-Factoid & 51.96 & 57.61 & 49.51 \\ 
 & BioASQ-List & 47.57 & 46.99 & 35.59 \\ 
 & BioASQ-Yesno & 21.20 & 95.20 & 68.25 \\ 
 & PubMedQA & 31.69 & 31.31 & 29.58 \\ 
\cmidrule(lr){1-5}
\multirow{3}{*}{\shortstack{NER}}  
 & AnatEM & \ \ 9.63 & 21.25 & 84.61 \\ 
 & BioNLP13CG & 26.03 & 41.16 & 65.09 \\ 
 & NCBI & 27.88 & 46.54 & 80.91 \\ 
\cmidrule(lr){1-5}
RE 
 & DDI & 51.00 & 53.40 & 89.35 \\ 
\cmidrule(lr){1-5}
SA 
 & Medical Drugs & 35.00 & 65.80 & 47.37 \\ 
\cmidrule(lr){1-5}
DC 
 & HoC & 62.84 & 62.65 & 82.53 \\ 
\midrule
\multicolumn{2}{c}{Average} & 36.48 & 52.19 & 63.23 \\ 
\bottomrule
\end{tabular}
\end{adjustbox}
\caption{Performance comparison (\textsc{\(F_1\)-Score}) with a fully supervised state-of-the-art instruction-tuned model \citep{parmar2022boxbart} in biomedicine, in which more than 140K training samples are involved.}
\label{table:SOTA}
\end{table*}

\begin{table*}[t!]
\centering
\begin{adjustbox}{width=1.0\textwidth} 
\begin{tabular}{llccc}
\toprule
\textsc{Biomedicine} & \textbf{Dataset} & \textbf{Base} & \textbf{Self-Specialized} & \textbf{AdaptLLM} \\ 
\midrule\midrule
\multirow{4}{*}{\shortstack{QA}} 
 & BioASQ-Factoid & 39.21 & 52.17 & 51.79 \\ 
 & BioASQ-List & 32.45 & 43.99 & 49.74 \\ 
 & BioASQ-Yesno & 66.00 & 88.40 & 93.80 \\ 
 & PubMedQA & 23.59 & 31.04 & 24.06 \\ 
\cmidrule(lr){1-5}
\multirow{3}{*}{\shortstack{NER}}  
 & AnatEM & \ \ 1.20	& 20.93 & 9.81 \\ 
 & BioNLP13CG & 22.16 & 31.46 & 18.58 \\ 
 & NCBI & 37.91 & 43.28 & 20.50 \\ 
\cmidrule(lr){1-5}
RE 
 & DDI & 47.6 & 53.80 & 51.00 \\ 
\cmidrule(lr){1-5}
SA 
 & Medical Drugs & 42.80 & 24.20 & 13.60 \\ 
\cmidrule(lr){1-5}
DC 
 & HoC & 10.60 & 50.87 & 8.33 \\ 
\midrule
\multicolumn{2}{c}{Average} & 32.35 & 44.01 & 34.12 \\ 
\bottomrule
\end{tabular}

\begin{tabular}{llccc}
\toprule
\textsc{Finance} & \textbf{Dataset} & \textbf{Base} & \textbf{Self-Specialized} & \textbf{AdaptLLM} \\ 
\midrule\midrule
\multirow{1}{*}{\shortstack{SUM}}
 & EDT-Summarization & 13.87 & 23.56 & 35.95 \\ 
\cmidrule(lr){1-5}
\multirow{2}{*}{\shortstack{QA}} 
 & ConvFinQA & 28.77 & 20.88 & 12.10 \\
 & InsuranceQA & 9.96 & 24.36 & 26.11 \\
\cmidrule(lr){1-5}
\multirow{2}{*}{\shortstack{NER}}  
 & Fin3 & 6.80 & 43.87 & 13.32 \\
 & FiNER\_139 & 44.34 & 35.63 & 11.10 \\
\cmidrule(lr){1-5}
\multirow{1}{*}{\shortstack{RE}}  
 & KPI-EDGER & 49.46 & 63.90 & 34.53 \\ 
\cmidrule(lr){1-5}
\multirow{3}{*}{\shortstack{SA}}  
 & EarningsCall & 49.03 & 47.74 & 47.30 \\
 & Financial\_Phrasebank & 29.20 & 68.80 & 34.80 \\
 & FIQA-SA & 61.54 & 70.09 & 69.66 \\ 
\cmidrule(lr){1-5}
\multirow{1}{*}{\shortstack{CLS}}  
 & Gold Commodity News & 38.42 & 61.20 & 71.61 \\ 
 \midrule
\multicolumn{2}{c}{Average} & 33.14 & 46.00 & 35.65 \\
\bottomrule
\end{tabular}
\end{adjustbox}
\caption{Comparison with a concurrent work, AdaptLLM \citep{cheng2024adapting}.}
\label{table:adaptllm}
\end{table*}

\begin{table*}[t!]
\centering
\begin{adjustbox}{width=0.8\textwidth} 
\begin{tabular}{lccc}
\toprule
\textbf{Models} & \textbf{\# Labeled Data} & \textbf{\# Synthetic Data} & \textbf{MMLU Score} \\ 
\midrule\midrule
Humpback-65B (General) & 3200 & 500K & 59.0 \\
Ours-7B (Specialized) & 100 & 5K & 64.0 \\
\bottomrule
\end{tabular}
\end{adjustbox}
\caption{Comparison with a concurrent work, Humpback \citep{li2024selfalignment}.}
\label{table:humpback}
\end{table*}

\section{Additioanl Results \& Discussion}
\label{sec:additional_results}
\subsection{Qualitative Analyses}
\label{subsec:qualitative_anslyses}
\begin{figure*}[!t]
    \centering
    \includegraphics[width=0.5\textwidth]
    {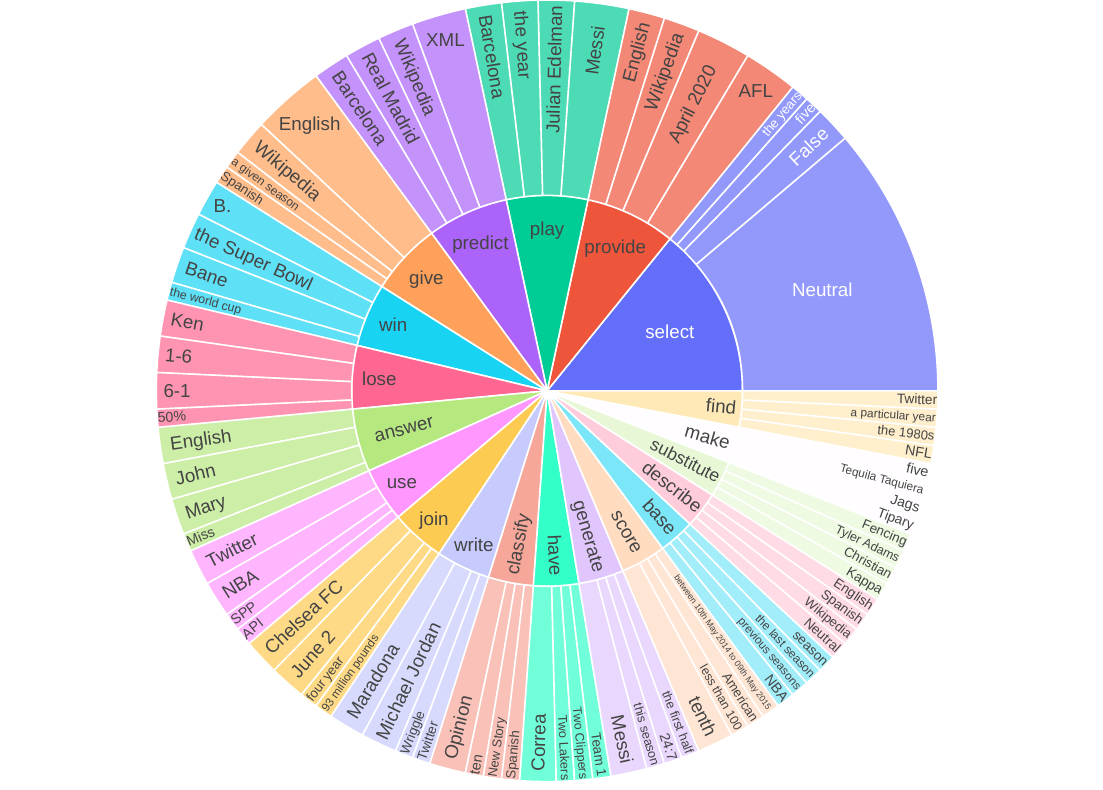}
    \includegraphics[width=0.37\textwidth]
    {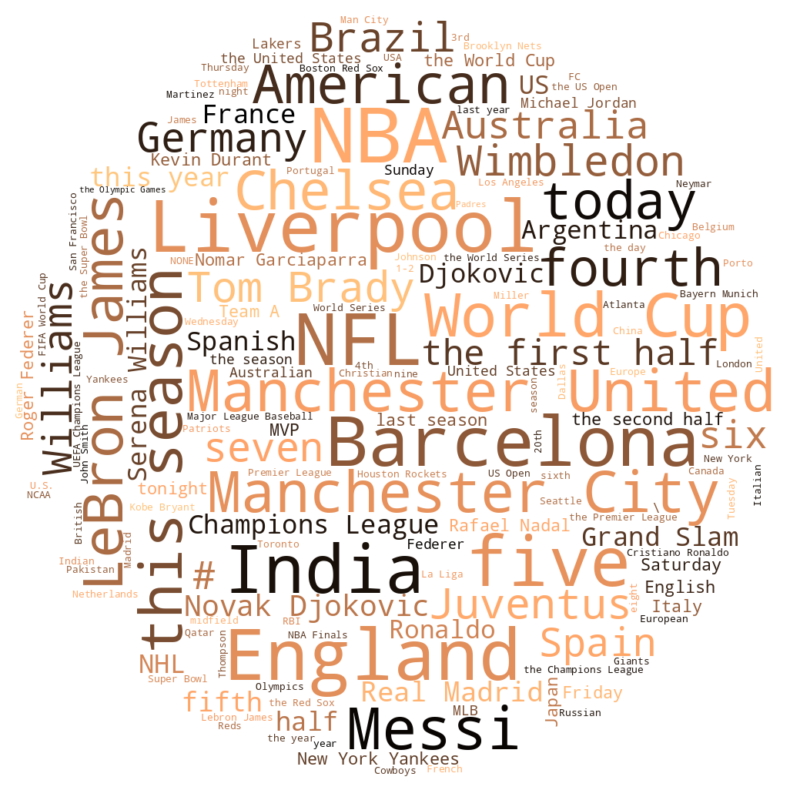}
    \caption{Statistics for instructions (left) and input context (right) generated through self-specialization toward the sports domain, with 40 seeds, 1st iteration only, and no retrieval component. On the left, the inner circle illustrates prevalent verbs in the instructions, with the outer ring revealing associated entities. Conversely, the right side showcases the input context, highlighting the diverse sports keywords generated by the model in the process of self-specialization. Best viewed in zoom and color.}
    \label{fig:stats_sports}
\end{figure*}

While our study primarily focuses on the biomedical and finance domain, the applicability and effectiveness of self-specialization in another specialized domain whose knowledge is relatively limited, such as sports, remain an open avenue for exploration. As an initial effort, we present a case study of a self-specialized model on sports in Table \ref{table:case_study_positive} \& \ref{table:case_study_negative}, along with the visualization of generated data in Figure \ref{fig:stats_sports}. We hope that this could offer insights into the versatility of self-specialization, although the model is not yet perfect, and thorough evaluations are required in future work. Different domains inherently pose unique requirements and nuances, and understanding how self-specialization adapts to these variations is a valuable direction for future work.

\subsection{On the Sensitivity of Prompting}
\label{subsec:sensitivity_of_prompting}
In Table \ref{table:main_results}, we observe the decreased performances with increased demonstrations in certain cases such as BioASQ and Medical Drugs. We conjecture this can be attributed to the model's sensitivity \citep{Zhao2021CalibrateBU} or interference among demonstrations \citep{Chen_2023} under in-context learning (ICL). In fact, it can even be noticed in the original GPT-3 paper \citep{NEURIPS2020_1457c0d6} that additional demonstrations do not always lead to better performance and can indeed sometimes result in a notable decrease, demonstrating an inherent challenge in ICL. Taking the worst, average, and the best across different k-shot (0, 1, 5) configurations for each dataset to address the concern of sensitivity, we still notice the significant gaps between our self-specialization and the base model, presented in Table \ref{table:main_results_best}.

\subsection{On Evaluation Designs}
\label{subsec:evaluation_desings}
In our study, as described in Section \ref{subsec:setups}, we treat all tasks as a unified text generation problem, aiming to assess the realistic capabilities of following instructions, consistent with established practices in biomedical instruction tuning literature \citep{parmar2022boxbart}. 
As briefly discussed in Section \ref{subsec:main_result}, we observe that in Table \ref{table:main_results}, the base model's performance on BioASQ-Yesno is very low (below random), often failing to follow instructions and generating text that is not confined to the label space. We therefore treat this dataset as an outlier and exclude it from our average calculations.  Even after removing this outlier, self-specialization still has substantial gains over the base model: 25.58 to 31.22 (0-shot), 28.42 to 36.55 (1-shot), and 32.55 to 43.21 (5-shot).
However, we believe that our current evaluation is fairer and preferable, because in a realistic scenario where a user prompts a model to solve a certain task (e.g., classification) without the assumption about a task type, and gets a totally wrong response out of the label space, evaluating such a response as correct would not make sense.

The primary objective of our work is to enhance the base model's domain-specific capabilities through self-specialization, a process inherently different from conventional fine-tuning approaches. Although the process utilizes LoRA for specialization, it is important to note that our approach fundamentally relies on synthetic data generated by the model itself. This unique aspect sets our method apart, as it effectively starts from scratch, focusing on self-generated, domain-specific instructional data for low-data scenarios. Finally, the base model and the base model improved through our Self-Specialization (using synthetic self-generated data) are compared fairly in the same zero-shot/few-shot setting.

\subsection{Comparison with State-of-the-art}
\label{subsec:SOTA_performances}
In Table \ref{table:SOTA}, we present the performances of a state-of-the-art instruction-tuned model \citep{parmar2022boxbart} in a biomedical domain, for a reference point. It is important to clarify that our comparison should not be considered direct. The SOTA model, unlike ours which relies on a few seed samples, is fine-tuned on a vast corpus of human-annotated data (140K), and differences in test set splits may exist. For the base MPT and self-specialized models, the maximum performances by using up to 5 samples are presented.

Despite significant improvements over its base model, our self-specialized model remains behind SOTA benchmarks, which is not surprising due to the nature of our method that is not supervised, unlike the SOTA model. While expected, this possibly implies the practical utility of our approach may be limited yet in certain scenarios.
From the table, we especially note the substantial gap in Named Entity Recognition (NER) tasks. This gap can be attributed to the SOTA model's training on a large and diverse set of NER datasets (i.e., 80K samples). This suggests ample opportunity for further exploration and enhancement in this area.

Additionally, we provide comparisons with concurrent works \citep{cheng2024adapting,li2024selfalignment} in Table \ref{table:adaptllm} and \ref{table:humpback}. To make distinctions, those works are, in principle, oriented toward different scenarios from ours. \citet{cheng2024adapting} leverages a large specialized corpus to transform them into reading comprehension texts using pre-defined templates, which are then used for training LLMs. Differently, our work does not necessitate such a huge amount of corpus that requires 768 GPU hours for training; ours takes only a few GPU hours. Moreover, we do not confine our method to specific tasks, aiming for cross-task generalization as shown in Table \ref{table:adaptllm}, unlike them focusing on constrained reading comprehension format.

\citet{li2024selfalignment} explores the augmentation of instructional data through the use of a few thousand labeled seed data to initialize their generation pipeline relying on a web corpus. Unlike them assuming huge amounts of seed/unlabeled data, our method requires only a handful of seeds (e.g., <100). Furthermore, their focus is on general instructions, which have shown only marginal effects in specialized domains in our preliminary results (Section \ref{sec:prelim}). Given the infeasibility of direct comparison, we reference their MMLU result from the paper to contextualize and we self-specialize Gemma-7B \citep{team2024gemma} on MMLU \citep{hendrycks2021measuring}. While acknowledging their values in different scenarios from ours, we believe this distinction provides an insight into the unique value and impact of our work in enhancing model performance along with efficiency in targeted domains.

\begin{figure*}[!t]
    \centering
    \includegraphics[width=1.0\textwidth]
    {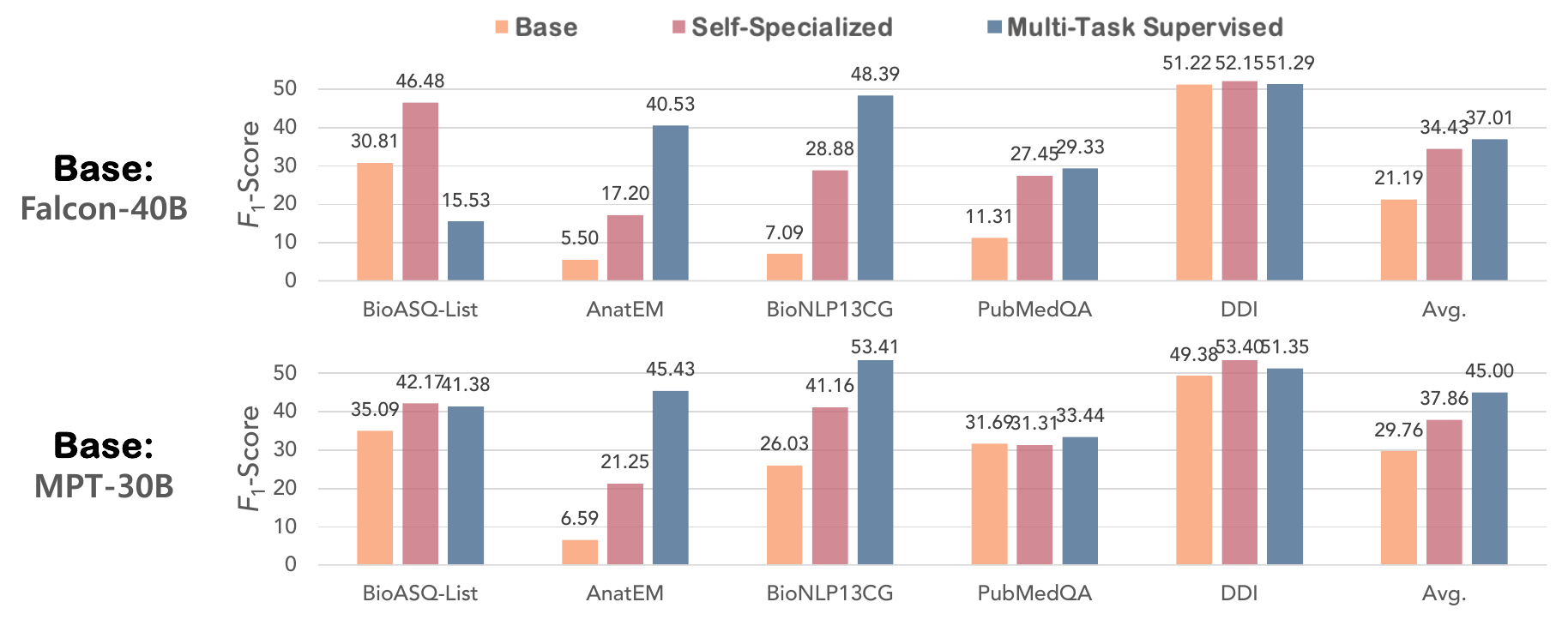}
    \caption{5-shot results based on Falcon-40B and MPT-30B, showcasing the self-specialization gains. ``Multi-Task Supervised" is a model trained on a large amount of human-labeled data in a multi-task setting and is provided \textit{for reference} as a (non-data-efficient, expensive) \textit{upper bound}.}
    \label{fig:falcon}
\end{figure*}

\begin{table*}[ht!]
\begin{center}
\begin{adjustbox}{width=1.0\textwidth}
\begin{tabular}{m{12cm}}
\toprule
\small{Instruction Generation Prompt} \\
\midrule
\tiny{\texttt{You are asked to come up with a set of 20 diverse task instructions about a biomedical domain. These task instructions will be given to a GPT model and we will evaluate the GPT model for completing the instructions. \newline\newline
Here are the requirements: \newline
1. Try not to repeat the verb for each instruction to maximize diversity. \newline
2. The language used for the instruction also should be diverse. For example, you should combine questions with imperative instructions. \newline
3. The type of instructions should be diverse. The list should include diverse types of tasks like open-ended generation, classification, editing, etc. \newline
4. A GPT language model should be able to complete the instruction. For example, do not ask the assistant to create any visual or audio output. For another example, do not ask the assistant to wake you up at 5pm or set a reminder because it cannot perform any action. \newline
5. The instructions should be in English. \newline
6. The instructions should be 1 to 2 sentences long. Either an imperative sentence or a question is permitted. \newline
7. You should generate an appropriate input to the instruction. The input field should contain a specific example provided for the instruction. It should involve realistic data and should not contain simple placeholders. The input should provide substantial content to make the instruction challenging. \newline
8. Ensure diverse tasks are covered in the instructions and inputs, while focusing on a biomedical domain. \newline\newline
List of 20 tasks:
}} \\

\midrule\midrule
\small{Response Generation Prompt} \\
\midrule

\tiny{\texttt{You are a biomedical domain expert. Given an instruction and an input, generate the best response to solve the given biomedical task.}} \\
\bottomrule
\end{tabular}
\end{adjustbox}
\end{center}
\caption{Prompts used for instruction generation and response generation.}
\label{table:prompts}
\end{table*}

\begin{table*}[ht!]
\begin{center}
\begin{adjustbox}{width=0.9\textwidth}
\begin{tabular}{m{12cm}}
\toprule
\tiny{\texttt{\textit{\underline{Instruction}} \newline
  The aim of this task is to identify the functional categories and chemical categories associated with a data record. The record is given in a pre-defined form and followed by a set of supporting sentences that might be relevant. In particular, for each data record, a triplet of functional categories, molecular categories and chemical categories associated with the data record will be provided. The task is to find some relations between records and supporting evidence that can support the given relations.
}} \\
\tiny{\texttt{\textit{\underline{Input}} \newline
Data record: Atom, CNA, NOC, NFT \newline
Text: Cisplatin and carboplatin, two drugs that are used to treat testicular and ovarian cancers, are not effective in treating breast cancer
}} \\
\tiny{\texttt{\textit{\underline{Output}} \newline
The functional category of cisplatin is CNA. The chemical category of cisplatin is NOC. The molecular category of cisplatin is NFT.
}} \\

\cmidrule(lr){1-1}

\tiny{\texttt{\textit{\underline{Instruction}} \newline
  In this task, you are given a short article and question. Read the short article and answer the question.
}} \\
\tiny{\texttt{\textit{\underline{Input}} \newline
Short article: The extract from the opium poppy was tested on breast cancer cells and was found to inhibit the migration and invasion of breast cancer cells. \newline
Question: Was the extract more potent in its inhibitory effect on the migration of breast cancer cells than its effect on the invasion of breast cancer cells?
}} \\
\tiny{\texttt{\textit{\underline{Output}} \newline
Yes.
}} \\

\cmidrule(lr){1-1}

\tiny{\texttt{\textit{\underline{Instruction}} \newline
  In this task, you are given a paragraph, and your task is to categorize the entities from the given input into two types: Dyes and Protein Molecules. The defined types here is based on the medical specialty such that dyes are typically for understanding human metabolism.
}} \\
\tiny{\texttt{\textit{\underline{Input}} \newline
Paragraph: Thus in an effort to improve the characterization of such compounds an alternative glycosylation labeling strategy was implemented based on malonate attachment to glucosamine and sond glycosyltransferase inactivation.
}} \\
\tiny{\texttt{\textit{\underline{Output}} \newline
Dyes: malonate, Protein Molecules: glucosamine, sond glycosyltransferase.
}} \\

\cmidrule(lr){1-1}

\tiny{\texttt{\textit{\underline{Instruction}} \newline
  In this task, you are given a small paragraph, your task is to identify the relationship between the variables. Find the most likely to be correct relationship of a gene with its characteristics. For example, if one entity is pathway, then you need to find the type of relationship that specific gene is related to that pathway. Give the most appropriate relationship type and tokens separating two variables. Given an input, determine the relationship type.
}} \\
\tiny{\texttt{\textit{\underline{Input}} \newline
COPD is a major cause of morbidity and mortality throughout the world, with over three million deaths per year Pathway: respiratory system Pathway: stress Hsa-miR-30b-5p is a regulator of retinal ganglion cell apoptosis through regulating the PTEN/AKT signaling pathway in glaucoma.
}} \\
\tiny{\texttt{\textit{\underline{Output}} \newline
Hsa-miR-30b-5p is a regulator of retinal ganglion cell apoptosis through regulating the PTEN/AKT signaling pathway in glaucoma.
}} \\

\cmidrule(lr){1-1}

\tiny{\texttt{\textit{\underline{Instruction}} \newline
  In this task, you are given a question and a sentence as context. You need to find the answers to the question in the given context. You can either answer the question or ask a question. Given the question, a context or a question, this task must either answer the question or ask a question.
}} \\
\tiny{\texttt{\textit{\underline{Input}} \newline
Question: For what drugs can predict the effect of changing the dose \newline
Context: Metoprolol is a commonly used beta-blocker, which competitively blocks beta-adrenergic receptors in the myocardium.  Adrenergic beta-blockers do not antagonize the presynaptic beta-1 receptors in the conduction system.  While they thus prolong the ventricular conduction time, they do not alter conduction velocity.  Metoprolol is therapeutically active over the heart rate range of 40 to 100 bpm, and may be particularly effective for the rate control of atrial fibrillation.  The beta-blocking action of metoprolol is accompanied by an increase in stroke volume.
}} \\
\tiny{\texttt{\textit{\underline{Output}} \newline
The beta-blocking action of metoprolol is accompanied by an increase in stroke volume.
}} \\

\bottomrule
\end{tabular}
\end{adjustbox}
\end{center}
\caption{A list of examples (\#1) of generated data on biomedicine through self-specialization.}
\label{table:examples1}
\end{table*}

\begin{table*}[ht!]
\begin{center}
\begin{adjustbox}{width=0.9\textwidth}
\begin{tabular}{m{12cm}}
\toprule
\tiny{\texttt{\textit{\underline{Instruction}} \newline
  In this task, you are given a drug name, and the side effect caused due to this drug (e.g., nausea). Your job is to generate other side effects in which this drug often causes.
}} \\
\tiny{\texttt{\textit{\underline{Input}} \newline
Drug\_1: Phentermine Drug\_2: Nausea
}} \\
\tiny{\texttt{\textit{\underline{Output}} \newline
Headache, insomnia, constipation, dry mouth, diarrhea, anorexia, increased blood pressure, increased heart rate, increased appetite, dizziness, nervousness, restlessness, irritability, tremor, rash, itching, and urticaria.
}} \\

\cmidrule(lr){1-1}

\tiny{\texttt{\textit{\underline{Instruction}} \newline
  Given a premise and a hypothesis sentence, your task is to identify if the premise supports the hypothesis or the premise contradicts the hypothesis sentence. Your task is to identify if the evidence in the premise supports or contradicts the hypothesis. There is no other alternative. You can select two labels for this task: Support or Contradict.
}} \\
\tiny{\texttt{\textit{\underline{Input}} \newline
Premise: If too much calcium enters the sea, if causes it to become more acidic in a process called "ocean acidification". Hypothesis: The ocean will become acidic.
}} \\
\tiny{\texttt{\textit{\underline{Output}} \newline
The vitamin did not seem to work for malaria. It was noted that the protein has is able to contact the insulin-like growth factor - 1 receptor, so an antagonist was made using a modified version of the protein.
}} \\

\cmidrule(lr){1-1}

\tiny{\texttt{\textit{\underline{Instruction}} \newline
  In this task, given the title of scientific study, your task is to generate a paper abstract or summary based on the given information.
}} \\
\tiny{\texttt{\textit{\underline{Input}} \newline
Title of study "Type and impact of end-user requests for application programming interfaces in three health-related internet services"
}} \\
\tiny{\texttt{\textit{\underline{Output}} \newline
The study found that the majority of end-user requests for application programming interfaces (APIs) in three health-related internet services were for data access. The study also found that the most common type of request was for data access, followed by data analysis and data visualization. The study also found that the most common impact
}} \\

\cmidrule(lr){1-1}

\tiny{\texttt{\textit{\underline{Instruction}} \newline
  In this task, you are given a question as input, and your task is to give answer it by generating a list of drugs which can be used for the treatment of the symptom. If the symptom is rare, the answer list should contain its corresponding differential diagnosis. Given a question related to symptoms, generate a list of drugs and differential diagnosis from open vocabulary.
}} \\
\tiny{\texttt{\textit{\underline{Input}} \newline
Question: What are the signs and symptoms of Leptospirosis?
}} \\
\tiny{\texttt{\textit{\underline{Output}} \newline
Fever, headache, muscle aches, nausea, vomiting, abdominal pain, diarrhea, cough, and red eyes. Differential diagnosis: Leptospirosis, Lyme disease, Rocky Mountain spotted fever, and dengue fever.
}} \\

\cmidrule(lr){1-1}

\tiny{\texttt{\textit{\underline{Instruction}} \newline
  In this task, you are given a list of paragraphs. The list contains sentences related to a particular biomedical concept and sentences not related to a particular biomedical concept. Classify these sentences as true or false. Answer if the sentences in the input are true or false. True indicates that all the sentences of the list are true for the given biomedical concept. False indicates that only one of the given sentences is false, while the rest of them are true.
}} \\
\tiny{\texttt{\textit{\underline{Input}} \newline
Some of the complications associated with developing small molecules for FGFR1 and KRAS targetable therapeutic compounds are revealed as the targets have low single amino acid sequence identity between species and both are, at least partially, regulated by post-translational modifications. S. Kisor, S. Khankorn, "Drug discovery from big data: past, present, and future.", Outline, January 2013.
}} \\
\tiny{\texttt{\textit{\underline{Output}} \newline
False
}} \\

\bottomrule
\end{tabular}
\end{adjustbox}
\end{center}
\caption{A list of examples (\#2) of generated data on biomedicine through self-specialization.}
\label{table:examples2}
\end{table*}

\newcolumntype{C}[1]{>{\centering\arraybackslash}m{#1}}
\begin{table*}[ht!]
\begin{center}
\begin{adjustbox}{width=1.0\textwidth}
\begin{tabular}{C{2.5cm} | C{6cm} | C{6cm}}
\toprule
 \textbf{Model} & \textbf{Base MPT-30B} & \textbf{Self-Specialized MPT-30B} \\
  \midrule\midrule
\textbf{Instruction} & \multicolumn{2}{c}
{\parbox{12cm}{\texttt{\scriptsize{Given a live commentary for a sports game, generate plausible sports news highlights.}}}} \\
\midrule
\textbf{Input Context} & 
    \multicolumn{2}{c}{\parbox{12cm}{
    \texttt{\scriptsize{Commentary \newline
    1. Full-time at the Etihad. Manchester City JUST manage to claim all three points in this Premier League clash as they overcome Swansea with a narrow 2-1 scoreline. The man of the match was Gabriel Jesus, who netted two goals, including the dramatic winner in the 92nd minute, which will prove to be a massive blow for the Swans. \newline
    2. GOOOOAAAALLLL! GOOOAAALLL FOR MANCHESTER CITY! ... \newline
    ...
    }}}} \\
\midrule
\textbf{Response} &
\multicolumn{1}{c|}{
\parbox{6cm}{
\texttt{\scriptsize{
A: I think you should try to use a pre-trained model for the encoder and decoder. \newline
...
}
}
}
}
& 
\multicolumn{1}{c}{
\parbox{6cm}{
\texttt{\scriptsize{
News: \newline
- Manchester City have come from behind to beat Swansea 2-1 in a thrilling Premier League clash at the Etihad Stadium. \newline
...
}
}
}
}
\\

\midrule\midrule

\textbf{Instruction} & \multicolumn{2}{c}
{\parbox{12cm}{\texttt{\scriptsize{This task aims to predict the sentiment of a tweet about sports. Given a tweet about football written by a real user, select an answer (polarity) among "Positive”, "Negative” and "Neutral”.}}}} \\
\midrule
\textbf{Input Context} & 
    \multicolumn{2}{c}{\parbox{12cm}{
    \texttt{\scriptsize{@user1 @user2 @user3 @user4 @user5 Now, PSG is considered a weak team. Poor ignorant. When Barça lost 3-0, it was because we faced PSG without Neymar. But now that Real Madrid eliminated them, suddenly PSG is a weak team. ...
    ...}}
    }} \\
\midrule
\textbf{Response} &
\texttt{\scriptsize{Positive}}
& 
\texttt{\scriptsize{Negative}}
\\

\midrule\midrule

\textbf{Instruction} & \multicolumn{2}{c}
{\parbox{12cm}{\texttt{\scriptsize{In this task, you are given a sentence related to sports. Your job is to determine whether the provided sentence makes sense. You need to produce an answer as "plausible” or "implausible”.}}}} \\
\midrule
\textbf{Input Context} & 
    \multicolumn{2}{c}{\parbox{12cm}{
    \texttt{\scriptsize{Lionel Messi hit a three-run homer.}}
    }} \\
\midrule
\textbf{Response} &
\texttt{\scriptsize{Plausible}}
& 
\texttt{\scriptsize{Implausible}}
\\

\midrule\midrule

\textbf{Instruction} & \multicolumn{2}{c}
{\parbox{12cm}{\texttt{\scriptsize{In this task, you are given a question and a context about sports, and you need to answer the question based on the given context.}}}} \\
\midrule
\textbf{Input Context} & 
    \multicolumn{2}{c}{\parbox{12cm}{
    \texttt{\scriptsize{
    Passage (Live Commentary): \newline
    As the match reaches a decisive point, Williams, the tennis star, finds herself engaged in a fierce rally. Her opponent delivers a powerful cross-court forehand that seems destined for a winner. However, Williams displays incredible reflexes and anticipation, executing a lightning-fast backhand passing shot down the line that catches her opponent off guard and secures the point. \newline
    Question: What remarkable shot did the player execute to win the crucial point? \newline
    Answer Choices: \newline
    a) Williams delivered a drop shot to surprise her opponent. \newline
    b) Williams hit a powerful forehand winner cross-court. \newline
    c) Williams executed a precise backhand passing shot. \newline
    d) Williams served an ace to win the point.
    }}}} \\
\midrule
\textbf{Response} &
\texttt{\scriptsize{a) Williams delivered a drop shot to surprise her opponent.}}
& 
\texttt{\scriptsize{c) Williams executed a precise backhand passing shot.}}
\\

\midrule\midrule

\textbf{Instruction} & \multicolumn{2}{c}
{\parbox{12cm}{\texttt{\scriptsize{In this task, you are given a question about sports. Answer the question based on your implicit knowledge.}}}} \\
\midrule
\textbf{Input Context} & 
    \multicolumn{2}{c}{\parbox{12cm}{
    \texttt{\scriptsize{What does the term "love" mean in tennis scoring?}}
    }} \\
\midrule
\textbf{Response} &
\texttt{\scriptsize{It means that the player has won the game.}}
& 
\texttt{\scriptsize{Love is a score of zero in tennis.}}
\\
\bottomrule
\end{tabular}
\end{adjustbox}
\end{center}
\caption{Case study on sports domain. Self-Specialized model produces better responses given the same instruction about sports.
}
\label{table:case_study_positive}
\end{table*}

\begin{table*}[ht!]
\begin{center}
\begin{adjustbox}{width=1.0\textwidth}
\begin{tabular}{C{2.5cm} | C{6cm} | C{6cm}}
\toprule
 \textbf{Model} & \textbf{Base MPT-30B} & \textbf{Self-Specialized MPT-30B} \\
  \midrule\midrule
\textbf{Instruction} & \multicolumn{2}{c}
{\parbox{12cm}{\texttt{\scriptsize{In this task, you are given a set of attribute-value pairs about sports in the raw XML format. Generate a natural language text describing the given data.}}}} \\
\midrule
\textbf{Input Context} & 
    \multicolumn{2}{c}{\parbox{12cm}{
    \texttt{\scriptsize{$<$originaltripleset$>$ \newline
        $<$otriple$>$batterName $|$ Pujols$<$/otriple$>$ \newline
        $<$otriple$>$finalScore $|$ 16-7$<$/otriple$>$ \newline
        $<$otriple$>$gameTally $|$ 2-1$<$/otriple$>$ \newline
        $<$otriple$>$hasLostTeam $|$ Texas$\_$Rangers$<$/otriple$>$ \newline
        $<$otriple$>$hasWonTeam $|$ St.$\_$Louis$\_$Cardinals$<$/otriple$>$ \newline
        $<$otriple$>$hitNumber $|$ five$<$/otriple$>$ \newline
        $<$otriple$>$homeRunNumber $|$ three$<$/otriple$>$ \newline
        $<$otriple$>$matchDate $|$ Saturday$\_$night$<$/otriple$>$ \newline
        $<$otriple$>$pitchResult $|$ homers$<$/otriple$>$ \newline
        $<$otriple$>$runNumber $|$ six$<$/otriple$>$ \newline
    $<$/originaltripleset$>$
    }}
    }} \\
\midrule
\textbf{Response} &
\texttt{\scriptsize{Albert Pujols hit three home runs and drove in five runs as the St. Louis Cardinals beat the Texas Rangers 16-7 on Saturday night.}}
& 
\texttt{\scriptsize{Pujols hit five home runs and drove in six runs as the St. Louis Cardinals beat the Texas Rangers 16-7 on Saturday night.}}
\\

\midrule\midrule

\textbf{Instruction} & \multicolumn{2}{c}
{\parbox{12cm}{\texttt{\scriptsize{In this task, you are given a sentence related to sports. Your job is to determine whether the provided sentence makes sense. You need to produce an answer as "plausible” or "implausible”.}}}} \\
\midrule
\textbf{Input Context} & 
    \multicolumn{2}{c}{\parbox{12cm}{
    \texttt{\scriptsize{Mike Trout took ball four in the World Series.}}
    }} \\
\midrule
\textbf{Response} &
\texttt{\scriptsize{Implausible}}
& 
\texttt{\scriptsize{Implausible}}
\\

\midrule\midrule

\textbf{Instruction} & \multicolumn{2}{c}
{\parbox{12cm}{\texttt{\scriptsize{In this task, you are given a question and a context about sports, and you need to answer the question based on the given context.}}}} \\
\midrule
\textbf{Input Context} & 
    \multicolumn{2}{c}{\parbox{12cm}{
    \texttt{\scriptsize{
    Passage (Live Commentary): \newline
    With two runners on base and a full count, Johnson, the opposing team's batter, faced an intense battle against the pitcher. The tension reached its peak as the pitcher delivered a devastating curveball, catching Johnson off guard. He swung and missed, resulting in a resounding strikeout that ended the inning and stranded the runners. \newline
    Question: When did the pitcher deliver a crucial strikeout to end the inning? \newline
    Answer Choices: \newline
    a) At the start of the inning, Johnson struck out. \newline
    b) After a series of foul balls, Johnson hit a double. \newline
    c) At the end of the inning, Johnson grounded out. \newline
    d) With a full count, Johnson struck out to end the inning.
    }}
    }} \\
\midrule
\textbf{Response} &
\texttt{\scriptsize{a) At the start of the inning, Johnson struck out.}}
& 
\texttt{\scriptsize{c) At the end of the inning, Johnson grounded out.}}
\\

\bottomrule
\end{tabular}
\end{adjustbox}
\end{center}
\caption{Case study on sports domain. Negative cases where both models produce wrong responses are presented.
}
\label{table:case_study_negative}
\end{table*}

\end{document}